%% file: main.tex
\documentclass[10pt,twocolumn,letterpaper]{article}

\usepackage[pagenumbers]{cvpr} 
\usepackage{bbm}
\usepackage{graphicx}
\usepackage{paralist}
\usepackage{amsmath}
\usepackage{amssymb}
\usepackage{booktabs}
\usepackage{xcolor}

\usepackage{multirow}

\usepackage{xcolor}

\newcommand{\blue}[1]{\textcolor{blue}{#1}}
\newcommand{\red}[1]{\textcolor{red}{#1}}

\usepackage[pagebackref,breaklinks,colorlinks]{hyperref}

\usepackage[capitalize]{cleveref}
\crefname{section}{Sec.}{Secs.}
\Crefname{section}{Section}{Sections}
\Crefname{table}{Table}{Tables}
\crefname{table}{Tab.}{Tabs.}

\def\cvprPaperID{8236} 
\def\confName{CVPR}
\def\confYear{2023}

\begin{document}

\title{Semi-supervised Parametric Real-world Image Harmonization}
\author{%
Ke Wang$^{1,2}$, Michaël Gharbi$^{1}$, He Zhang$^{1}$, Zhihao Xia$^{1}$, Eli Shechtman$^{1}$\\
{$^1$ Adobe Inc.}\\
{$^2$ EECS, University of California, Berkeley}\\
{\tt\small\{kewang, mgharbi, hezhan, zxia, elishe\}@adobe.com}\\
{\tt\small kewang@berkeley.edu}\\
}

\twocolumn[{%
\renewcommand\twocolumn[1][]{#1}%
\renewcommand\twocolumn[1][]{#1}%
\maketitle
\begin{center}
    \centering
    \captionsetup{type=figure}
\vspace{-8mm}
    \includegraphics[width=0.95\linewidth]{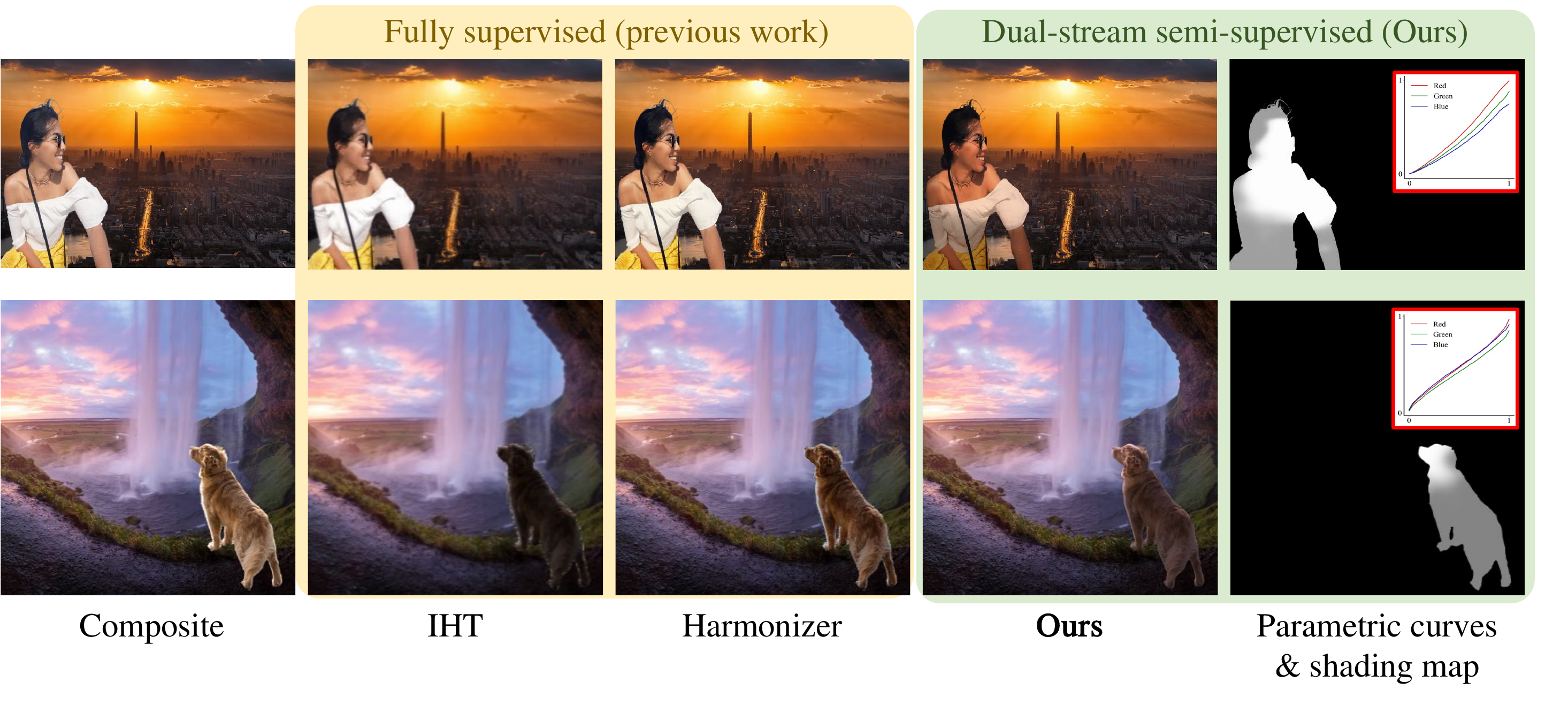}
\vspace{-4mm}
    \captionof{figure}{\label{fig:teaser}
        \textbf{Visual comparisons between state-of-the-art harmonization methods IHT~\cite{guo2021image}, Harmonizer~\cite{ke2022harmonizer}, and ours.}
        Our  model is fully parametric.  This gives artists full posterior control over the final composite, makes runtime efficient for high-resolution real-world inputs and regularizes training.
        Our model predicts \textit{global RGB curves} and a \textit{local shading map} (right).
        Benefiting from the novel dual-stream semi-supervised training strategy, our method (right) produces more realistic harmonized images on real-world composites (left).
        This new training strategy, together with the shading map, makes it the first harmonization method to address local tonal adjustments, such as shading the face according to the sun's direction (top) or selectively darkening the part of the dog inside the cave (bottom).
    }
\end{center}%
}]

\input{TexFolder/abstract.tex}


\input{TexFolder/introduction.tex}

\input{TexFolder/Related_Works.tex}

\input{TexFolder/method.tex}

\input{TexFolder/Experiments.tex}

\input{TexFolder/Conclusion.tex}
\clearpage
\newpage

{\small
\bibliographystyle{ieee_fullname}
\bibliography{egbib}
}

\clearpage
\newpage

\input{TexFolder/Supplementary_og.tex}

\end{document}

%% file: TexFolder/abstract.tex
\begin{abstract}
\vspace{-.12in}
%
\noindent
Learning-based image harmonization techniques are usually trained to undo synthetic random global transformations applied to a masked foreground in a single ground truth photo.
This simulated data does not model many of the important appearance mismatches (illumination, object boundaries, etc.) between foreground and background in real composites, leading to models that do not generalize well and cannot model complex local changes.
We propose a new semi-supervised training strategy that addresses this problem and lets us learn complex local appearance harmonization from unpaired real composites, where foreground and background come from different images.
Our model is fully parametric.
It uses RGB curves to correct the global colors and tone and a shading map to model local variations.
Our method outperforms previous work on established benchmarks and real composites, as shown in a user study, and processes high-resolution images interactively.

\end{abstract}

%% file: TexFolder/introduction.tex
\vspace{-5mm}

\section{Introduction}

%
Image harmonization~\cite{reinhard2001color,pitie2005n,jia2006drag,sunkavalli2010multi,xue2012understanding,tao2010error} aims to iron out visual inconsistencies created when compositing a foreground subject onto a background image that was captured under different conditions~\cite{xue2012understanding,niu2021making},
by altering the foreground's colors, tone, etc., to make the composite more realistic.
Despite significant progress, the practicality of today's most sophisticated learning-based image harmonization techniques~\cite{cong2020dovenet,cong2022high,guo2021image,guo2021intrinsic,jiang2021ssh,liang2021spatial,ke2022harmonizer,xue2012understanding} is limited by a severe domain gap between the synthetic data they are trained on and real-world composites.

%
As shown in Figure~\ref{fig:domain-gap}, the standard approach to generating synthetic training composites applies global transforms (color, brightness, contrast, etc.) to a masked foreground subject in a ground truth photo.
This is how the iHarmony Dataset~\cite{cong2020dovenet,fivek} was constructed.
%
A harmonization network is then trained to recover the ground truth image from the synthetic input.
While this approach makes supervised training possible, it is unsatisfying in simulating the real composite in that synthetic data does not simulate mismatch in illumination, shadows, shading, contacts, perspective, boundaries, and low-level image statistics like noise, lens blur, etc. However, in real-world composites, the foreground subject and the background are captured under different conditions, which can have more diverse and arbitrary differences in any aspects mentioned above.

\begin{figure}[t]
\hspace*{-0.5cm}
  \centering
   \quad \includegraphics[width=1\linewidth]{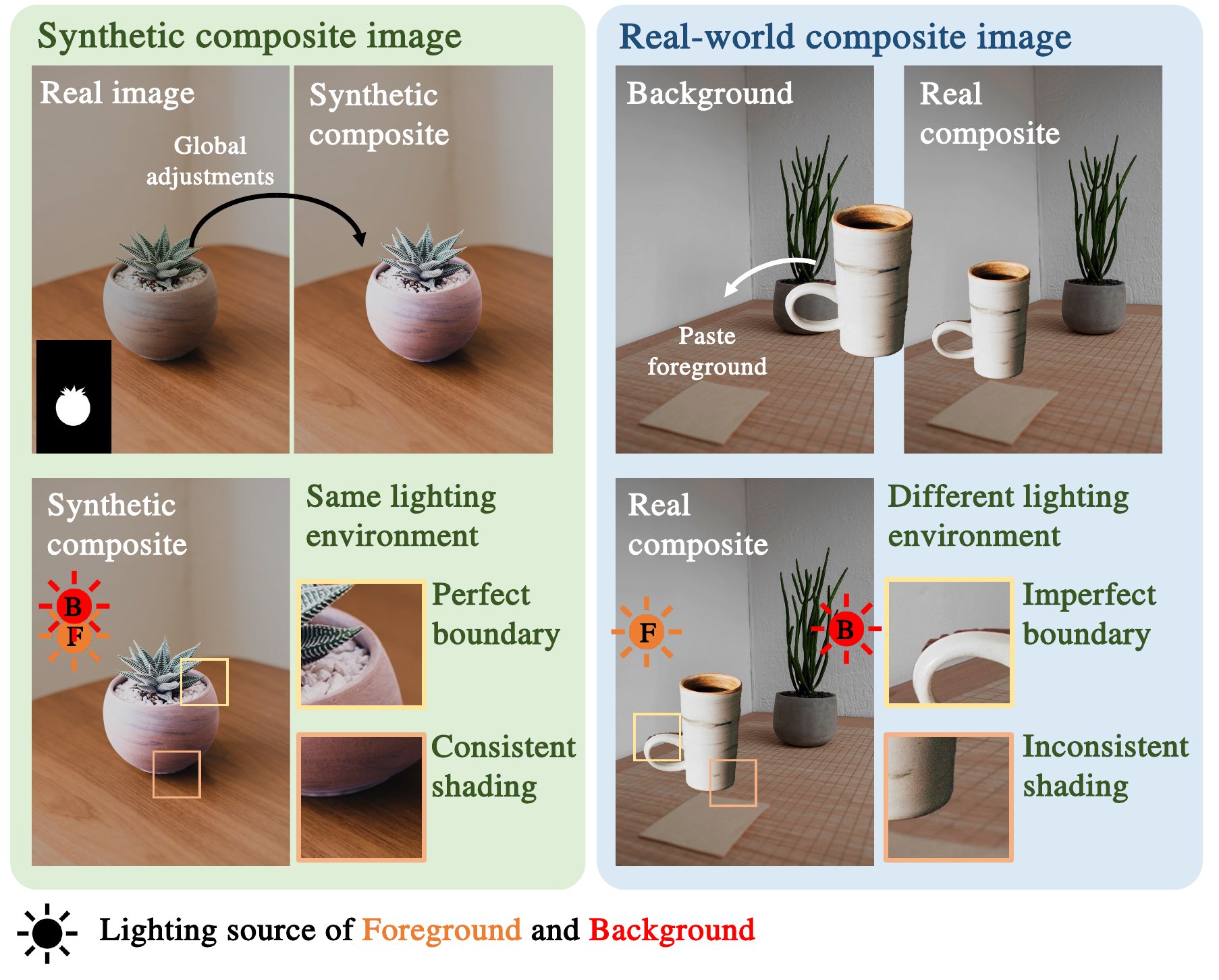}
    \vspace{-6mm}
   \caption{\label{fig:domain-gap} \textbf{Domain Gap between synthetic and real-world composites.}
   The existing synthetic composites \cite{cong2020dovenet} (left), generated by applying global transforms (e.g., color, brightness), are unable to simulate many of the appearance mismatches that occur in real composites (right).
   This leads to a domain gap: models trained on synthetic data do not generalize well to real composites.
   In real composites (right), the foreground and background are captured under different conditions.
   They have different illuminations, the shadows do not match, and the object's boundary is inconsistent.
   Such mismatches do not happen in the synthetic case (left). 
   }
   \label{fig:domain-gap}
  \vspace{-4mm}
\end{figure}

%
We argue that using realistic composites for training is essential for image harmonization to generalize better to real-world use cases. 
Because collecting a large dataset of artist-created before/after real composite pairs would be costly and cumbersome, our strategy is to use a semi-supervised approach instead.
We propose a novel dual-stream training scheme that alternates between two data streams. 
Similar to previous work, the first is a supervised training stream, but crucially, it uses artist-retouched image pairs.
Different from previous datasets, these artistic adjustments include global color editing but also dodge and burn shading corrections and other local edits.

%
The second stream is fully unsupervised.
It uses a GAN~\cite{goodfellow2020generative} training procedure, in which the critic compares our harmonized results with a large dataset of realistic image composites.
Adversarial training requires no paired ground truth.
The foreground and background for the composite in this dataset are extracted from different images so that their appearance mismatch is consistent with what the model would see at test time.
%

\begin{figure*}
\vspace{-1mm}
    \quad \includegraphics[width=0.95\textwidth]{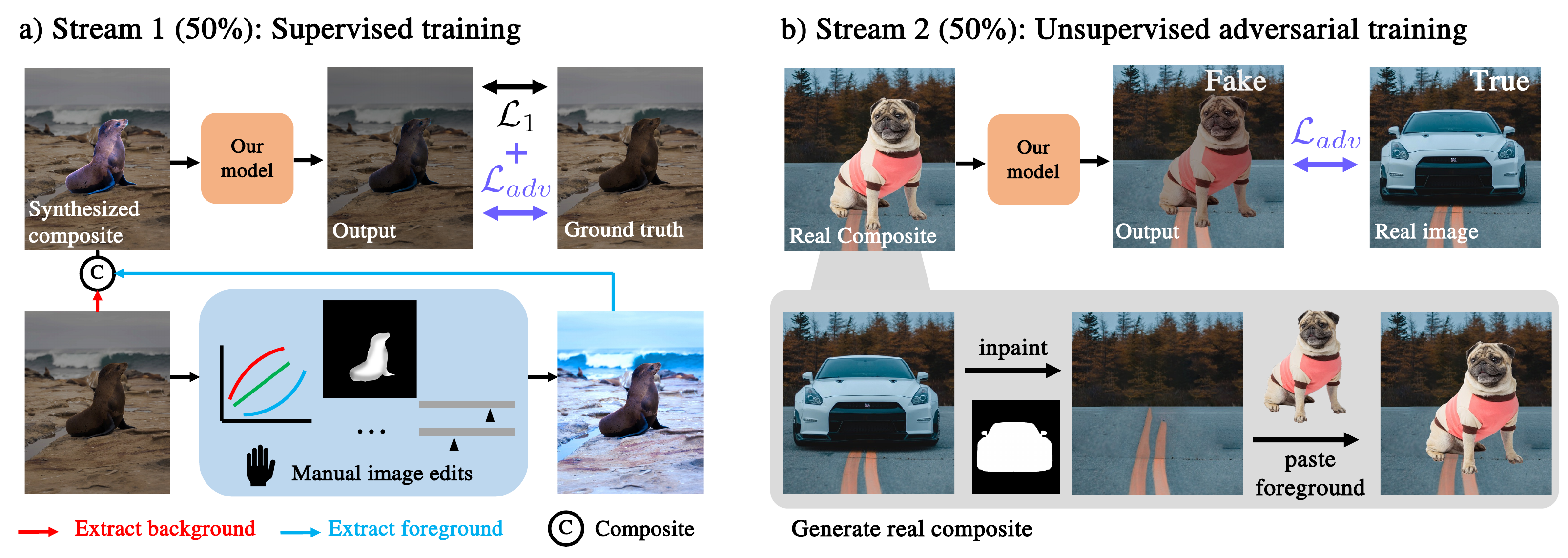}
    \vspace{-2mm}
    \caption{\label{fig:harmonization-framework} \textbf{Overview of semi-supervised dual-stream training strategy.} 
 To bridge the domain gap, our proposed semi-supervised dual-stream training strategy alternates between two training streams: a) Supervised training with artist-retouched composite image pairs  (left). Artist adjustments include global color editing, shading correction, and other local edits. b) Unsupervised adversarial training with real-world composite images (right). It uses a GAN\cite{goodfellow2020generative} training procedure, comparing our harmonized results with a large dataset of realistic image composites. The foreground and background for the composite are from different images, so the appearance mismatch is consistent with what we see at test time.}
\vspace{-2mm}
\end{figure*}
%
%
%

To reap the most benefits from our semi-supervised training, we also introduce a new model that is fully parametric.
To process a high-resolution input composite at test time, our proposed network first creates
a down-sampled copy of the image at $512\times 512$ resolution, from which it predicts \emph{global RGB curves} and a smooth, low-resolution \emph{shading map}.
We then apply the RGB curves pointwise to the high-resolution input and multiply them by the upsampled shading map.
The shading map enables more realistic local tonal variations, unlike previous harmonization methods limited to global tone and color changes, either by construction~\cite{ke2022harmonizer,xue2022dccf,liang2021spatial} or because of their training data~\cite{cong2020dovenet}.
%

%
Our parametric approach offers several benefits.
First, by restricting the model's output space, it regularizes the adversarial training. Unrestricted GAN generators often create spurious image artifacts or 
other unrealistic patterns~\cite{zhang2019detecting}.
Second, it exposes intuitive controls for an artist to adjust and customize the harmonization result post-hoc. This is unlike the black-box nature of
most current learning-based approaches~\cite{cong2020dovenet,cong2022high,guo2021image,guo2021intrinsic}, which output an image directly.
And, third our parametric model runs at an interactive rate, even on very high-resolution images (e.g., 4k), whereas several state-of-the-art methods~\cite{guo2021image,guo2021intrinsic,cong2020dovenet} are limited to low-resolution (e.g., $256\times 256$) inputs.

To summarize, we make the following contributions:
\vspace{-2mm}
\begin{itemize}
  \item A novel dual-stream semi-supervised training strategy that, for the first time, enables training from real composites, which contains much richer local appearance mismatches between foreground and background.
\vspace{-2mm}
 \item A parametric harmonization method that can capture these more complex, local effects (using our shading map) and produces more diverse and photorealistic harmonization results.

%
\vspace{-2mm}
  \item State-of-the-art results on both synthetic and real composite test sets in terms of quantitative results and visual comparisons, together with a new evaluation benchmark.
\end{itemize}

We will release our code publicly upon publication.

%% file: TexFolder/Related_Works.tex
\section{Related works}
\vspace{-2mm}
\noindent\textbf{Image harmonization}.
Traditional image harmonization methods mainly focus on adjusting the low-level appearance statistics (e.g., color statistics, gradient information) between the foreground objects and the background~\cite{reinhard2001color,pitie2005n,jia2006drag,sunkavalli2010multi,xue2012understanding,tao2010error}.
Supervised learning-based approaches have been proposed and shown notable success~\cite{cong2020dovenet,cong2022high,guo2021image,guo2021intrinsic,tsai2017deep,zhu2015learning} by learning image harmonization from synthetic training pairs, for instance, iHarmony Dataset\cite{cong2020dovenet}. Works as DIH\cite{tsai2017deep}, DovNet\cite{cong2020dovenet}, IHT\cite{guo2021image}, Guo \textit{et al.}\cite{guo2021intrinsic} consider the image harmonization task as a pixel-wise image-to-image translation task, and are limited to low-resolution inputs (typically $256\times256$) due to computational inefficiency. Recent work extended image harmonization to high-resolution images by designing parametric models \cite{cong2022high,liang2021spatial,ke2022harmonizer,xue2022dccf}. To name a few,  Liang \textit{et al.} learns the spatial-separated RGB curves for high-resolution image harmonization. Ke \textit{et al.}~\cite{ke2022harmonizer} directly predicts the filter arguments of several white-box filters, which can be efficiently applied to high-resolution images.
In all of those approaches, synthetic training pairs are generated by applying global transforms to the masked foreground subjects of natural ground-truth images and hence do not simulate mismatch in illumination, shadows, shading, contact, etc., that happen in real-world composite images. Therefore, due to the synthetic training data and model construction\cite{ke2022harmonizer,liang2021spatial}, previous works are limited to global tone and color changes. In contrast, our model is trained on real-world composite images and artist-retouched synthetic images, which enables us to model richer image edits and produce more compelling results on real composites.

\noindent\textbf{Efficient and high-resolution image enhancement.} There has been a wide range of research focusing on designing efficient and high-resolution image enhancement algorithms\cite{gharbi2015transform,gharbi2017deep,moran2020deeplpf}. 
Gharbi \textit{et al.}~\cite{gharbi2017deep} introduced a convolutional neural network (CNN) that predicts the coefficients of a locally-affine model in bilateral space from down-sampled input images. 
The coefficients are then mapped back to the full-resolution image space.
Zeng \textit{et al.}~\cite{zeng2020learning} directly learns 3D Lookup Tables (LUTs) for real-time image enhancement.
In our application, image harmonization can be considered as a background-guided image enhancement problem.
Thus, inspired by~\cite{gharbi2017deep, zeng2020learning}, we design a network that directly predicts the coefficients of RGB curves (piece-wise linear function) from down-sampled composite inputs. We then apply the RGB curves pointwise to the high-resolution input without introducing extra computation costs.

\noindent\textbf{Image-based relighting} 
Image-based relighting approaches \cite{pandey2021total,sun2019single,philip2021free, yeh2022learning} focus on modifying the input lighting conditions and local shading to generate convincing composite results. However, recent relighting methods are mainly designed for portrait applications and do not generalize well to other objects due to the limitation of Light-stage capturing only portraits but not diverse objects \cite{debevec2000acquiring}. With a similar idea of incorporating local shading edits but a different approach, our method embeds the shading layer into a network and trains on composite image datasets without explicitly leveraging scene representations (geometry, materials, lighting) and using full relighting models.


%% file: TexFolder/method.tex
\vspace{-2mm}
\section{Method}

Our image harmonization method corrects the foreground subject in a rough composite to make the overall image look more realistic using a new parametric model (\S~\ref{sec:parametric_harmonization}) that can be applied to real-world high-resolution images efficiently.
Previous harmonization techniques train on synthetically-generated
composite pairs~\cite{cong2020dovenet}, where the model's input is a global transformation of a ground truth image within a foreground subject mask.
%
%
%
The colors are often unnatural, the mask boundary is close to perfect, and there is no mismatch in appearance, illumination, or low-level image statistics since both foreground and background come from the same image.
As a result, models trained on such data generalize poorly.
Our method addresses this crucial issue using a novel dual-stream semi-supervised training strategy (\S~\ref{section:dual}) that leverages high-quality artist-created image pairs and unpaired realistic composites to bridge the training-testing domain gap.
See Figure~\ref{fig:harmonization-framework} for an overview.
%
%

\begin{figure}[t]
  \centering
\hspace*{-0.5cm}
  
   \includegraphics[width=1\linewidth]{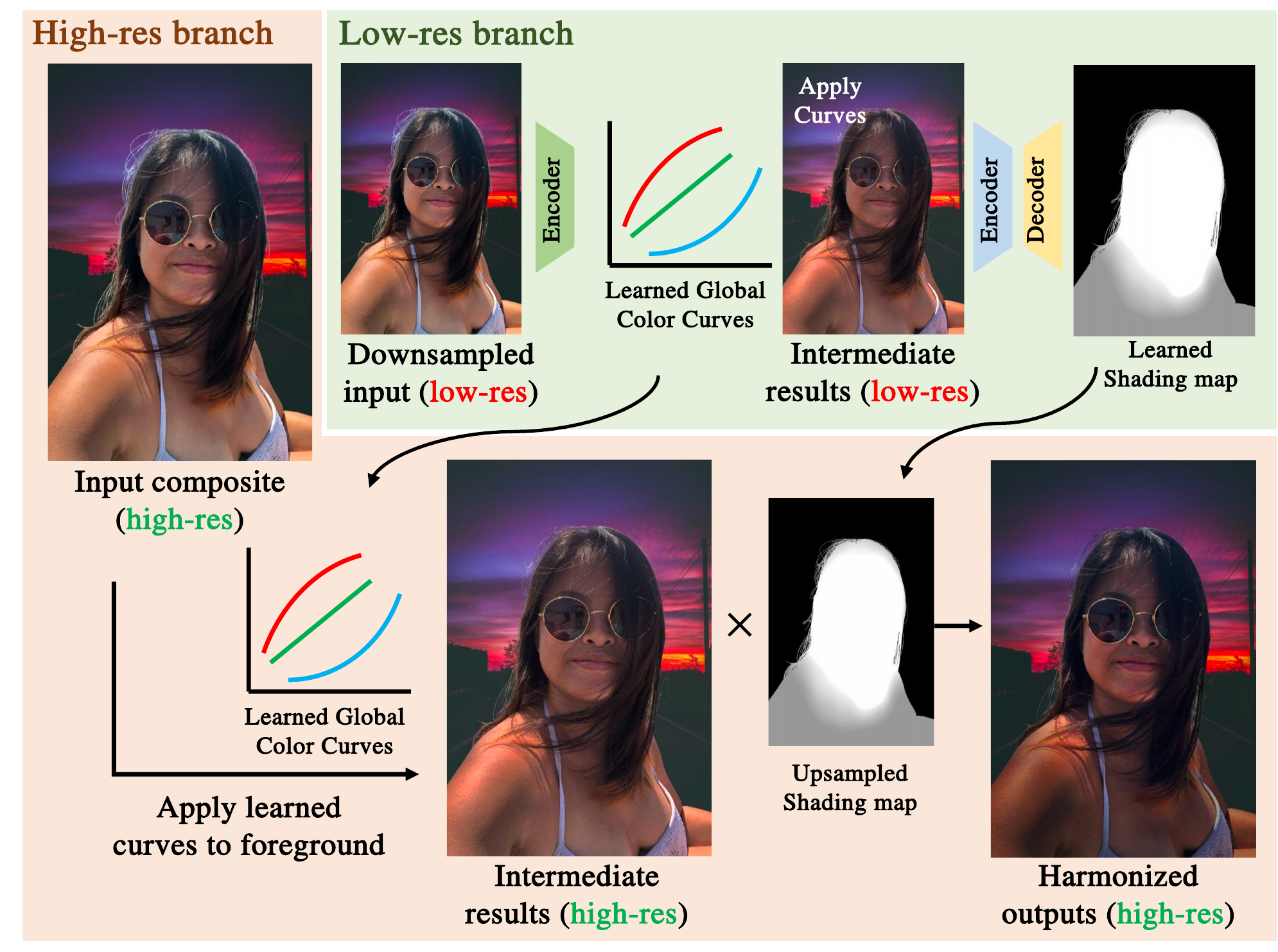}
   \caption{\label{fig:pih}\textbf{Illustration of our parametric model design.} Our framework consists of a low-resolution branch and a high-resolution branch. At test time, we down-sample the given high-resolution image and predict the \textit{global RGB curves and shading map} through a two-stage network. Those parametric outputs are then executed at the original resolution to produce the final harmonized image. Our model can scale to any resolution.}

  \vspace{-4mm}
   
\end{figure}

\subsection{Parametric image harmonization}\label{sec:parametric_harmonization}
Our network design is inspired by real-world composite harmonization workflows\footnote{\url{https://youtu.be/g3qe4rDw1XU}}.
An artist typically applies several image corrections sequentially, each dedicated to harmonizing a specific composite element, such as luminosity, color, or shading.
Accordingly, our image transformation model consists of two modules, applied sequentially: a pointwise global color correction using RGB curves and a local shading correction using a low-frequency multiplicative map.
%
For efficiency, our model operates at two resolutions.



%
\vspace{2mm}
\noindent\textbf{Pipeline overview.} As illustrated in Figure~\ref{fig:pih}, our harmonization pipeline takes as input a foreground image $\mathbf{F}\in\mathbb{R}^{3hw}$ with dimensions $h,w\in\mathbb{N}$, a background image image $\mathbf{B}\in\mathbb{R}^{3hw}$, and a compositing alpha mask $\mathbf{M}\in\mathbb{R}^{hw}$.
We define the \emph{unharmonized} composite image as $\mathbf{C} := \mathbf{M}\cdot\mathbf{F} + (1-\mathbf{M})\cdot\mathbf{B}$. 
At test time, we start by downsampling the inputs to a fixed resolution ($512\times512$ ), denoting the low-resolution images by $\mathbf{C}^{lr}$, $\mathbf{B}^{lr}$, $\mathbf{M}^{lr}$ respectively.
We concatenate these maps and pass them to a neural network $f$ that predicts the parameters $[\theta_1, \theta_2] :=f(\mathbf{C}^{lr},\mathbf{B}^{lr},\mathbf{M}^{lr})$ of our two-stage parametric image transformation.
Finally, we apply the parametric transformation $t_1$, $t_2$ sequentially on the high-resolution input to obtain the final harmonized composite $\mathbf{O} := t_2(t_1(\mathbf{C}, \mathbf{M};\theta_1), \mathbf{M};\theta_2)$, where $\mathbf{M}$ is used to ensure only the area under the foreground mask is altered.
We describe the two stages in the parametric transformation next.

%

%

%
%
%
\vspace{2mm}
\noindent\textbf{Global color correction curves.}
Our first high-resolution processing stage $t_1$ applies the predicted global RGB curves for color correction.
We parameterize them as 3 piecewise linear curves with 32 control points, applied to each color channel independently so that $\theta_1\in\mathbb{R}^{32\times2\times3}$ is the set of 2D coordinates of the curve nodes.
%
%
We predict these parameters from $[\mathbf{C}^{lr},\mathbf{B}^{lr}, \mathbf{M}^{lr}]$ using a ResNet-50\cite{he2016deep}-based network.
%
%
Applying the curve is a per-pixel operation that can be efficiently computed at any resolution.

\vspace{2mm}
\noindent\textbf{Local low-frequency shading map.}
Our second stage $t_2$ multiplies the image with a low-frequency grayscale shading map, to model local tonal corrections.
It is applied on the output of the first stage.
We constrain the shading map to only model low-frequency change by generating at a low resolution: $\theta_2\in\mathbb{R}^{64\times64}$.
It is produced by a modified U-Net\cite{ronneberger2015u} with large receptive field, given the low-resolution buffers $[\mathbf{C}^{lr},\mathbf{B}^{lr}, \mathbf{M}^{lr}]$, together with the output of the color-correction stage at low-resolution $t_1(\mathbf{C}^{lr}, \mathbf{M}^{lr};\theta_1)$
At test time, we upsample the low-resolution shading map to the original high-resolution and multiply it pointwise with the color-corrected image to obtain our 
final harmonized composite:
\vspace{-1.5mm}
\begin{equation}
    \mathbf{O} = t_1(\mathbf{C}, \mathbf{M}; \theta_1) \cdot \text{upsample}(\theta_2).
\vspace{-0.5mm}
\end{equation}

%
%
%

\subsection{Dual-stream semi-supervised training}\label{section:dual}

Our semi-supervised training strategy aims to alleviate the generalization issues that plague many state-of-the-art harmonization models, as shown in Figure~\ref{fig:domain-gap}.
Our approach uses two data streams, sampled with equal probability, and minimizes a different objective on each data stream.
The first stream uses input/output composite pairs similar to previous work, except that we only use artist-created image transformations instead of random augmentations. 
The second is unsupervised. This allows us to use more realistic images obtained by compositing foreground and background from unrelated images, for which no ground truth is easily obtainable. The objective for the supervised stream is the combination of $\ell_1$ loss and adversarial loss, while we only impose adversarial loss for the unsupervised training stream. 

%

\vspace{2mm}
\noindent\textbf{Supervised training using retouched images.} The first stream is fully supervised. Unlike previous work, we use images retouched by artists rather than mostly relying on random augmentations.
We refer to this dataset as \emph{Artist-Retouched} in the rest of the paper.
Artists were allowed to use common image editing operations such as global luminosity or color adjustments, but also local editing tools like brushes, e.g., to alter the shading.
%
Specifially, we collected $n=46173$ before/after retouching image pairs $\{\mathbf{I}_i, \mathbf{O}_i\}_{i=1,\ldots,n}$, with the mask for one foreground object $\mathbf{M}_i$ for each pair.
From each triplet, we can create 2 input composites for training: one with only the foreground retouched $\mathbf{M}_i\cdot\mathbf{O}_i + (1-\mathbf{M}_i)\cdot\mathbf{I}_i$, and the other with only the background is retouched $\mathbf{M}_i\cdot\mathbf{I}_i + (1-\mathbf{M}_i)\cdot\mathbf{O}_i$. 
Since our harmonization model only alters the foreground, we use the unedited image $\mathbf{I}_i$, and the retouched image $\mathbf{O}_i$ as ground truth targets for these input composites, respectively.

When sampling training data from this stream, we optimize our model's parameters to minimize the sum of an 
$\ell_1$ reconstruction error $\mathcal{L}_{rec}$ between the ground truth and our model output, and an adversarial objective~\cite{goodfellow2020generative}
%
%
%
\vspace{-1.5mm}
\begin{equation}
   \lambda \mathcal{L}_{rec} + (1-\lambda)\mathcal{L}_{\text{G}},
 \vspace{-1.5mm}
\end{equation}
with $\lambda$ balances the two losses. For our experiments, $\lambda$ is empirically set to 0.92.
The generator, our parametric image harmonization model, is trained to produce outputs that cannot be distinguished from “real” images.
We use a U-Net discriminator~\cite{wang2021real} $D$ to make per-pixel real vs.\ fake classifications.
Since our data formation model assumes the background is always correct, our discriminator is trained to predict
the inverted foreground mask $1-\mathbf{M}$.
That is when shown ``fake'' images, i.e., the background pixels have label 1 and the foreground 0.
For the ``real'' class, the target is all an all-1s map.
%
%
%
So the discriminator loss is given by:
\vspace{-1.5mm}
\begin{equation}
\begin{split}
    \mathcal{L}_{D} = &- \mathbb{E}_{\mathbf{I}_{real}}[\log(D(\mathbf{I}_{real}))] \\
    &- \mathbb{E}_{\mathbf{I}_{fake}}[\log((1-\mathbf{M}) - D(\mathbf{I}_{fake}))],
    \label{eq:d_loss}
\end{split}
\end{equation}
%
%
The generator loss is:
%
\begin{equation}
    \mathcal{L}_{G} = - \mathbb{E}_{\mathbf{I}_{fake}}[\log(D(\mathbf{I}_{fake}))].
    \label{eq:g_loss}
\vspace{-1.5mm}
\end{equation}
%

%

To further increase the training diversity, we randomly augment the foreground brightness on the fly without retouching the color.

\vspace{2mm}
\noindent\textbf{Unsupervised training with real composites.} Our second training stream is unsupervised.
It uses randomly generated composites that are representative of real-world use cases but for which no ground truth is available.
To properly reproduce the appearance mismatch in real applications, we create these composites as follows.
We start from a dataset of $m$ images $\{\mathbf{I}_i\}_{i=1,\ldots,m}$, each with a foreground object mask $\mathbf{M}_i$, from which we derive a foreground $\mathbf{F}_i = \mathbf{M}_i\cdot\mathbf{I}_i$ and a background $\mathbf{B}_i = (1-\mathbf{M}_i)\cdot\mathbf{I}_i$.
As preprocessing, we dilate the foreground mask by 30 pixels and inpaint the corresponding area in the background image using a pre-trained inpainting network (we use LaMa~\cite{suvorov2022resolution}).
Then during, training we sample two images $i$ and $j$ and create a composite by pasting the foreground $j$ onto the inpainted background of $i$:
%
\vspace{-1.5mm}
\begin{equation}
    \mathbf{C}_{ij} := \mathbf{F}_j\cdot\mathbf{M}_j + \text{inpaint}(\mathbf{B}_i, \mathbf{M}_i)\cdot(1-\mathbf{M}_j).
\vspace{-1.5mm}
\end{equation}
The triplet $[\mathbf{C}_{ij}, \text{inpaint}(\mathbf{B}_i, \mathbf{M}_i), \mathbf{M}_j]$ is passed as input to our model.
Figure~\ref{fig:harmonization-framework}b illustrates the process.

%
%
%

%
With no ground truth available when sampling composites from this data stream,
we only optimize the adversarial loss $(1-\lambda)\mathcal{L}_{\text{G}}$, as
defined in Eq.~\eqref{eq:g_loss}, where again the fake samples
$\mathbf{I}_{fake}$ are the outputs of our model.


The discriminator is trained with Eq.~\eqref{eq:d_loss}, where $\mathbf{I}_{real}$ is not a real composite, but is obtained by masking the foreground subject $\mathbf{F}_i$, inpainting the background $\mathbf{B}_i$, and pasting the foreground back onto the same image, i.e.
\vspace{-1.5mm}
\begin{equation}
    \mathbf{I}_{real} := \mathbf{F}_i\cdot\mathbf{M}_i + \text{inpaint}(\mathbf{B}_i, \mathbf{M}_i)\cdot(1-\mathbf{M}_i).
\vspace{-1.5mm}
\end{equation}
This is similar to how we produce a composite of two images $i$ and $j$, expect that we only use one image, $i$.
This alteration of the ``real'' class is to prevent the discriminator from using the inpainting boundary region as a strong cue to discriminate between our model output and real images, which leads to collapse in the GAN training.

%
%
%

GAN training is known to be unstable or cause image artifacts~\cite{zhang2019detecting}, but because our parametric harmonization model adjusts color curves and adds low-resolution shadows, instead of predicting pixels directly, it has a strong regularizing effect, which prevents the GAN training to degenerate and cause spurious artifacts in the output image.
We use the same discriminator (and generator) in both streams.

%

%% file: TexFolder/Experiments.tex
\section{Experiments}
\begin{figure*}[t]
\vspace{-6mm}

  \centering
   \includegraphics[width=1\linewidth]{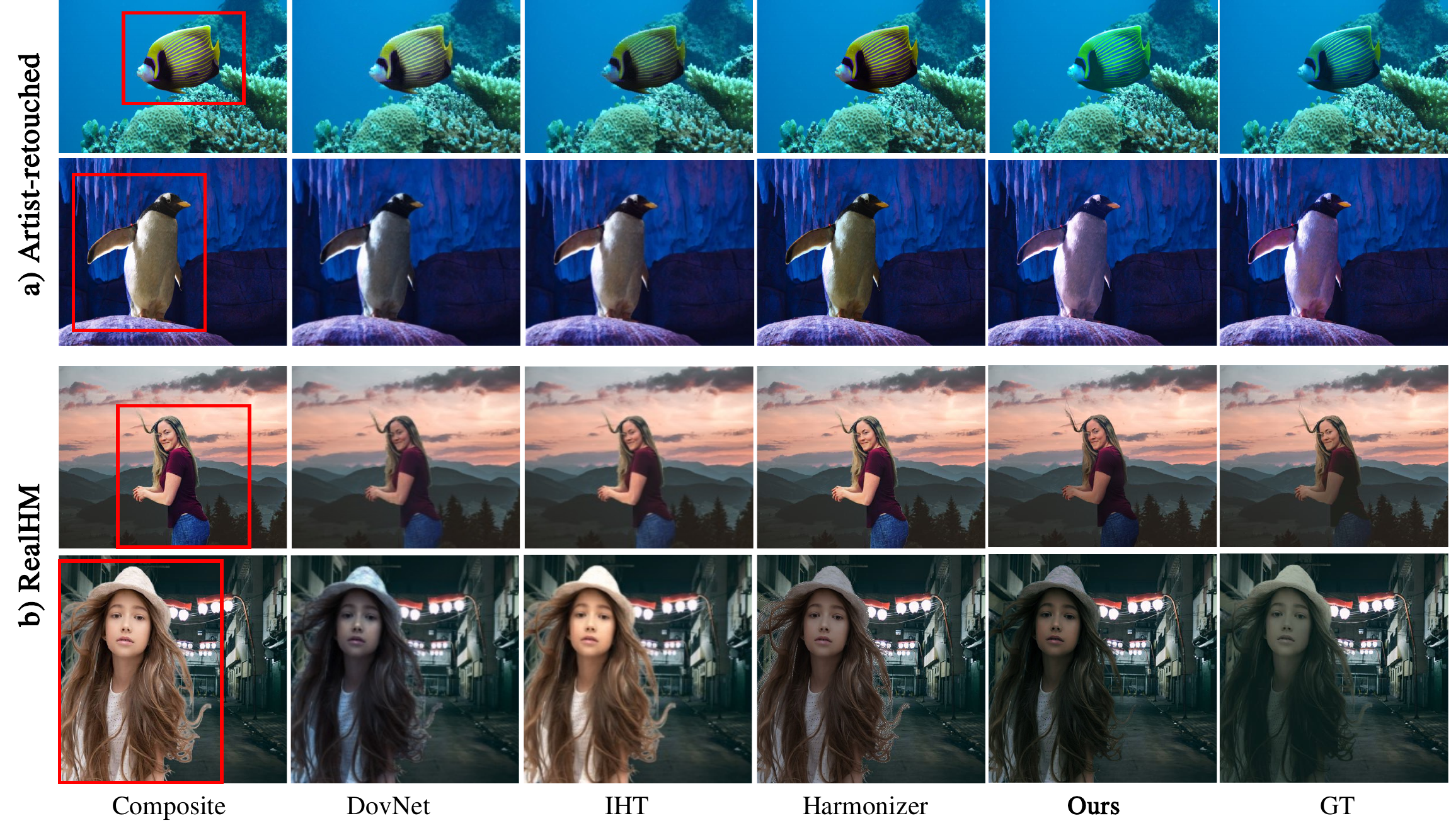}
\vspace{-6mm}
   
      \caption{\label{fig:visual_comparison}
         \textbf{Representative visual comparisons between state-of-the-art harmonization results.} We compared our method with composite, DovNet~\cite{cong2020dovenet}, IHT~\cite{guo2021image}, and Harmonizer~\cite{ke2022harmonizer}, and ground truth on both a) \emph{Artist-Retouched} synthetic dataset and b) RealHM real-world composite dataset. Red boxes indicate the foreground subject in the composite image. The ground truth for RealHM benchmark\cite{jiang2021ssh} is expert-annotated harmonization results. Our results show better visual agreements with the ground truth in terms of color harmonization (rows 1,2 and 4) and shading correction (row 3).
         }
\vspace{-6mm}
\end{figure*}
%
%
We compare our parametric image harmonization model with state-of-the-art methods on several established benchmarks (\S~\ref{sec:quant}), as well as a test subset of \emph{Artist-Retouched} dataset.
Furthermore, we demonstrate our superior performance on real-world harmonization tasks via a user study and qualitative comparisons on real composites (\S~\ref{section:user_study}).
Ablations of our method demonstrate the benefits of our semi-supervised training strategy and the individual components of our parametric model (\S~\ref{sec:abl}).
More results can be found in the supplementary.

\noindent\textbf{Evaluation metrics:} For quantitative comparisons with ground truth, we report performances by Mean Square Error (MSE), Peak Signal-to-Noise Ratio (PSNR), Structural Similarity (SSIM), and Learned Perceptual Image Patch Similarity (LPIPS)\cite{zhang2018unreasonable}. PSNR is measured in dB and calculated as: $\textrm{PSNR} = 10\log_{10}\frac{255^{2}}{\textrm{MSE}}$.

\noindent\textbf{Implementation details:} Our model and discriminator are implemented in PyTorch~\cite{paszke2019pytorch} and trained on
an NVIDIA A100 GPU using the Adam optimizer\cite{kingma2014adam} for 80 epochs, with a batch size of 8 and an initial learning rate of $4\times10^{-5}$, decayed by a factor $0.2$  every 20 epochs.

%
%
%

\subsection{Quantitative comparisons on paired data}\label{sec:quant}
We compare our method with three recent methods, DovNet~\cite{cong2020dovenet}, Image Harmonization with Transformer (IHT)~\cite{ke2022harmonizer}, and Harmonizer~\cite{ke2022harmonizer}, using the pre-trained model released publicly by the authors.
We first evaluate the synthetic iHarmony benchmark~\cite{cong2020dovenet}.
For fairness, our method uses the same setup as theirs for this comparison.
In particular, we train our model exclusively on the same training set as the baselines, using only our fully-supervised stream, deactivating the adversarial loss, and only passing the composite $\mathbf{C}$ and foreground mask $\mathbf{M}$ as inputs.
We report metrics at both at $256\times 256$ resolution and at $2048\times 2048$ on the HAdobe5k high-resolution subset of iHarmony.
Like our parametric approach, Harmonizer can process high-res images, but the other two methods are limited to $256\times256$ inputs.
So, for high-res comparison, we bilinearly downsample the input to DovNet and IHT, process the image, then bilinearly upsample the result before computing the metrics.
Despite its simplicity, our parametric model consistently outperforms or matches the more complex baselines.
Results are summarized in Table~\ref{table:iharm}.

\input{TexFolder/table_iharmony}

\input{TexFolder/table_cooper_rhm}

%
%
%
%
%
%

The iHarmony dataset is dominated by unrealistic synthetic image augmentations (71\%), so we also evaluate our results on more realistic retouches from
human experts. 
The two datasets we use for evaluation are a testing split of our \emph{Artist-Retouched} dataset, introduced in Section~\ref{section:dual}, containing 1000 before/after pairs, and the RealHM~\cite{jiang2021ssh} benchmark, containing 216 real-world high-resolution composites with expert annotated harmonization results as ground truth.
We compared the performance of their pre-trained models and ours trained with the full dual-stream pipeline at $2048\times2048$ resolution.
Table~\ref{table:realhm} shows our method consistently outperforms the baselines, with around 30\% relative MSE improvements compared to Harmonizer~\cite{ke2022harmonizer} on both datasets. 
As shown in Figure~\ref{fig:visual_comparison}, our method produces more realistic results, closer to the ground truth.

%
%


%
%
%
%
%

%
%

%
%
\subsection{Evaluation on real composite images}\label{section:user_study}

Our semi-supervised training procedure allows us to train on realistic composites, where foreground and background come from different sources.
Just like it limits the training potential of harmonization methods, using paired data created from a single ground truth image for evaluation is unsatisfying because it is not representative of real-world use cases (Fig.~\ref{fig:domain-gap}).
So, we demonstrate the practical effectiveness of our method in a user study with real composites.
For qualitative evaluation, we also created a set of 40 high-resolution real composite images with  reference images.

\noindent\textbf{User Study.}
%
Our user study follows a 2 alternatives forced choice protocol\cite{zhang2018unreasonable}, comparing our model with DovNet~\cite{cong2020dovenet}, IHT~\cite{guo2021image}, and Harmonizer~\cite{ke2022harmonizer}.
%
We selected 60 real composites from the RealHM dataset~\cite{jiang2021ssh}, making sure there were no duplicate foregrounds or backgrounds.
Since RealHM primarily focuses on portrait images, we also created 40 non-portrait real composites using free-to-use images  from Unsplash~\footnote{\url{https://unsplash.com/}}, giving us a total of 100 real composite images.
Each of our results is compared with the unaltered input composite and the three baseline results, which gives $100\times4 = 400$ image pairs to compare in total, which we submitted for evaluation to a pool of subjects on Amazon Turk~\footnote{\url{https://www.mturk.com/}}.
Each participant was shown 50 image pairs and, for each pair, they were asked to ``select which image looks more plausible''.
To ensure the quality of the responses, each subject was also shown  10 `sentinel' testing pairs composed of a real natural image and an extremely off-retouch image (e.g., where the image is all green).
This helped us filter low-quality participants, such as users that always click `left' to try and game the MTurk reward.
%
After filtering, we obtained pair-wise comparison results from 70 subjects, contributing a total of 3500 comparisons.
To analyze these results, we follow previous work~\cite{cong2020dovenet,cong2022high,ke2022harmonizer}, and use the Bradley-Terry (B-T)~\cite{bradley1952rank} model to derive the global ranking of all methods. 
We normalize the B-T scores such that the sum of the scores equals one across methods.
Table~\ref{table:btscore} summarizes the results.
It shows that our method achieves the highest B-T score, outperforming all the baselines, indicating our approach compares favorably in real-world applications.

\input{TexFolder/table_user_study}

\noindent\textbf{Real composites with captured reference.}
Figure~\ref{fig:real_captured_gt} shows two representative examples of real composite results (see supplemental for more).
For this qualitative comparison, we created a dataset of 40 high-resolution real-composite images with reference images by capturing a fixed set of foreground objects against multiple backgrounds, as well as a `background-only' image.
By segmenting the foreground object from one photo and pasting onto the `background-only' image of another, we get an input composite for our model.
The captured photo of the same object in the same background scene (placed at roughly the same location) acts as qualitative reference.
Compared to other approaches, our results are visually closer to the captured reference.
%
%
%

\begin{figure}[ht]
\hspace*{-0.3cm}
  \centering
\vspace{-2mm}
   \includegraphics[width=1\linewidth]{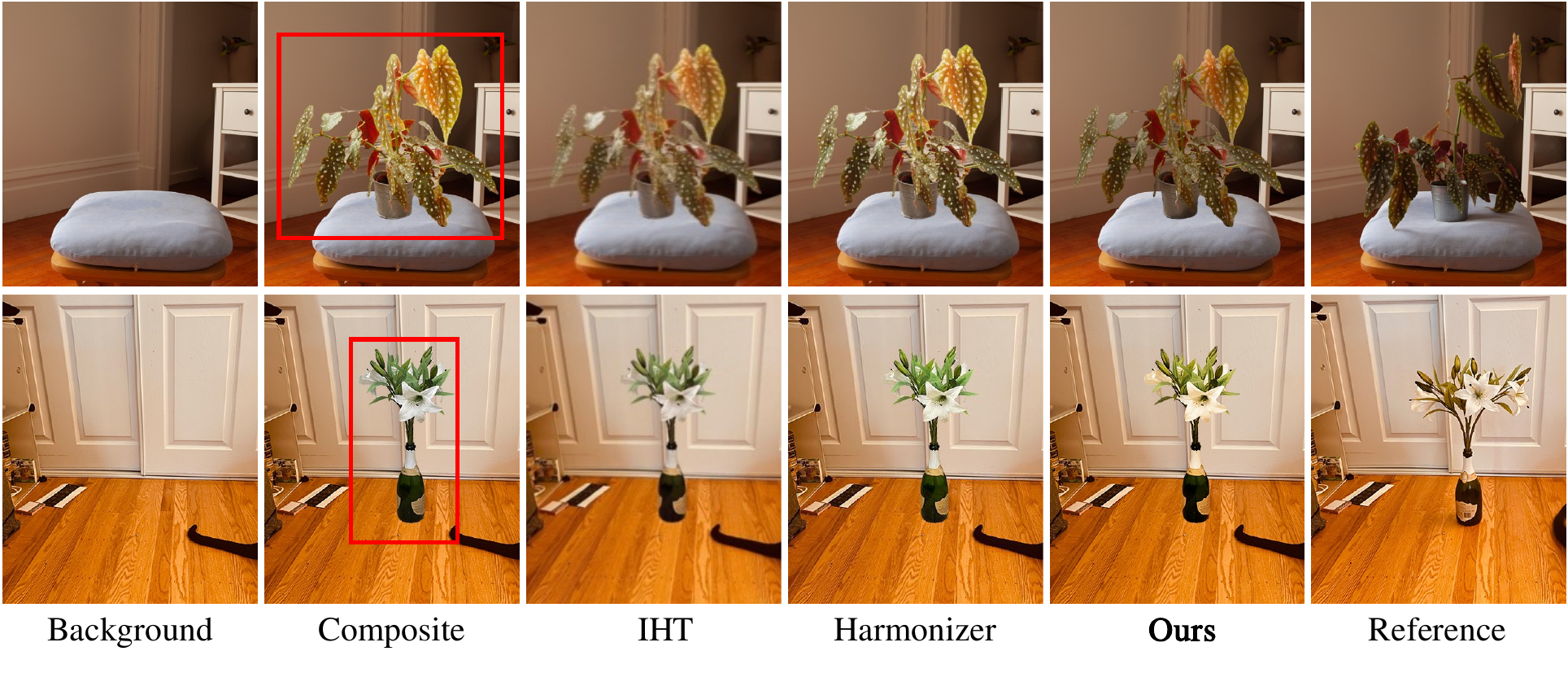}
   \caption{\label{fig:real_captured_gt}
   \textbf{Real composite harmonization results with captured reference.}
   The composite is obtained by pasting the foreground subject, from a different photo (not shown) onto the background (left).
   The reference (right) is obtained by physically placing the foreground subject in the background scene and taking a photo.
   We compare our method with IHT~\cite{guo2021image}, and Harmonizer~\cite{ke2022harmonizer}.
   Our results show better visual agreement with the captured reference (best viewed by zooming on the digital preprint).
   }
  \vspace{-4mm}
\end{figure}



\subsection{Ablation studies}\label{sec:abl}

\begin{figure}[ht]
\hspace*{-0.2cm}
  \centering
   \includegraphics[width=1\linewidth]{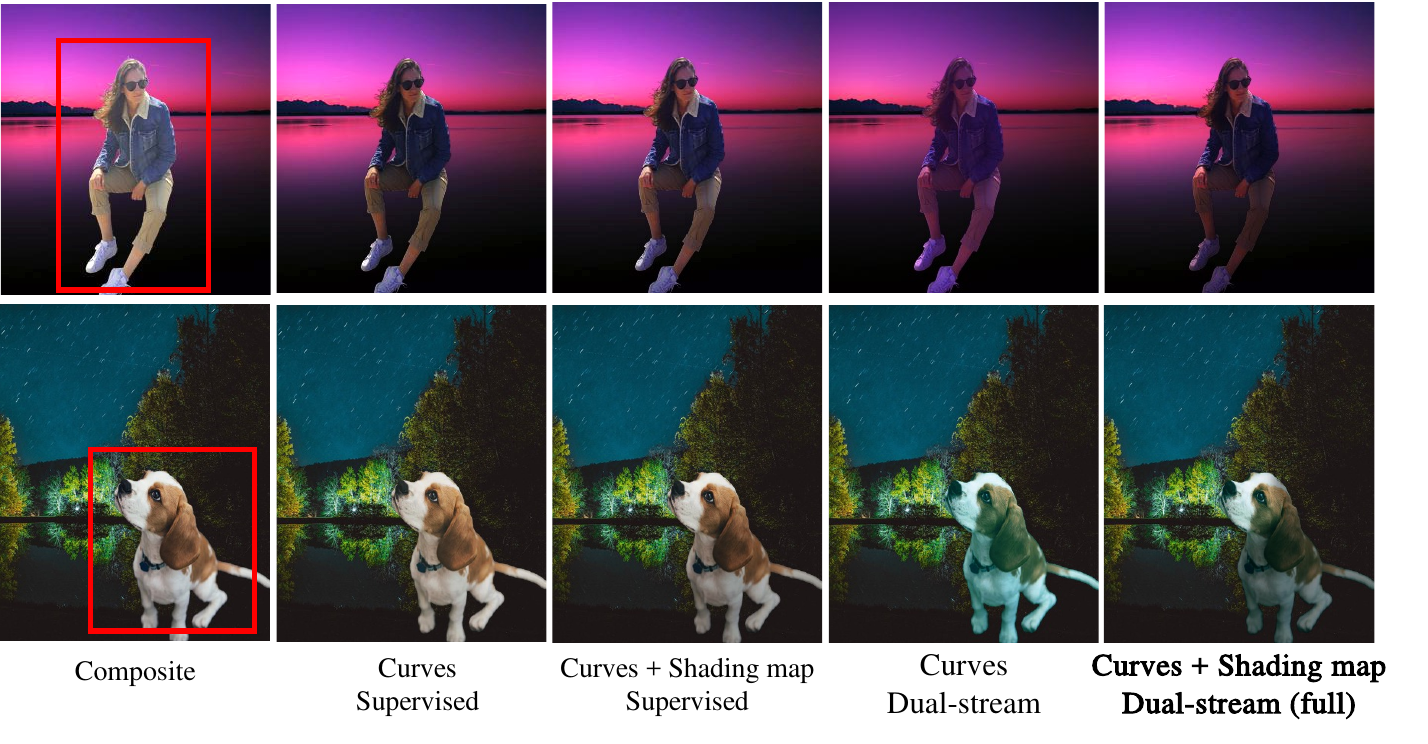}
  \vspace{-.15in}
   \caption{\label{fig:ablation}\textbf{Visual comparison of ablations.}
    Our full pipeline (right) shows more color-harmonious results than supervised training-only models (columns 2 and 3).
    Our local shading map adjusts local shading and produces more natural outputs (compare columns 4 and 5).
    }
  \vspace{-4mm}
\end{figure}

We evaluate the benefits of our dual-stream semi-supervised training strategy, comparing it to conventional supervised training.
We also analyze the impact of our global RGB curve module and shading map.
We conduct the comparisons on RealHM~\cite{jiang2021ssh} at $512\times 512$ resolution, comparing our full method (dual-stream training + two-stage model) with:
\begin{inparaenum}
\item Supervised training only (Stream 1) + global curves only;
\item Supervised training only (Stream 1) + two-stage parametric model;
\item Dual-stream training + global curves module only. 
\end{inparaenum}
We report quantitative metrics (MSE and PSNR), and the B-T score from a user study (similar to \S~\ref{section:user_study}, but this time with 68 subjects)
%
Table~\ref{table:ablation} and Figure~\ref{fig:ablation} summarize our results.
They show our shading map and our dual-stream training strategy both significantly improve realism over the curve-only, fully-supervised model.

\begin{table}[ht]
\centering
\setlength\tabcolsep{1.6pt} 

\begin{tabular}{ccccccc} 
\toprule
\begin{tabular}[c]{@{}c@{}}Stream \\1\end{tabular} & \begin{tabular}[c]{@{}c@{}}Stream\\~2\end{tabular} & \begin{tabular}[c]{@{}c@{}}Global\\Curves\end{tabular} & \begin{tabular}[c]{@{}c@{}}Shading\\Map\end{tabular} & MSE & PSNR & B-T score  \\ 
\midrule
                                                \checkmark  &   & \checkmark  &                  & 288.7     &  26.45    &        0.201    \\
                                                  \checkmark  &    &\checkmark  &  \checkmark & 264.0 &  26.89    &         0.206   \\
                                                  \checkmark & \checkmark &   \checkmark  & &   222.6  &  27.29    &  0.217         \\
                                                  \checkmark & \checkmark & \checkmark & \checkmark &  \red{\textbf{154.3}}   &  \red{\textbf{28.40}}  &    \red{\textbf{0.252}}        \\ 
\midrule
\multicolumn{4}{c}{Composite}                                                                                                                                                                                           & -   & -    &      0.124      \\
\bottomrule
\end{tabular}
\vspace{-2mm}
\caption{\textbf{Ablation study results of training strategies and parametric model.} We compare our semi-supervised training strategy (Stream 1 + Stream 2) with supervised training (Stream 1) and compare our two-stage model (Global Curves + Shading map) versus the model with only the global curve module. MSE and PSNR are used for quantitative comparisons, and the B-T score is calculated from user study results. }
\label{table:ablation}
\end{table}

As reported in Table \ref{table:ablation}, we observe that the dual-stream training strategy outperforms supervised training (row 3 and 4 \textit{v.s.} row 1 and 2) in terms of both quantitative metrics and B-T score, which demonstrates the benefits of our proposed dual-training strategy in real-world applications. Inspecting the results in Figure \ref{fig:ablation}, we observe that the dual-training strategy (column 4 and 5) brings advantages in color-harmonization when there is a strong foreground-background color mismatch.

On the other hand, as shown in Table \ref{table:ablation} row 3 \textit{v.s.} row 4, our proposed two-stage parametric model outperforms the global curve-only model by a large margin on RealHM benchmark, reducing the MSE by 30\%. Furthermore, as shown in Figure \ref{fig:ablation}, our full model (last column) includes both color harmonization and local shading to the results, achieving more plausible and harmonious results.


\begin{figure}[ht]
\hspace*{-0.4cm}
  \centering
   \includegraphics[width=\linewidth]{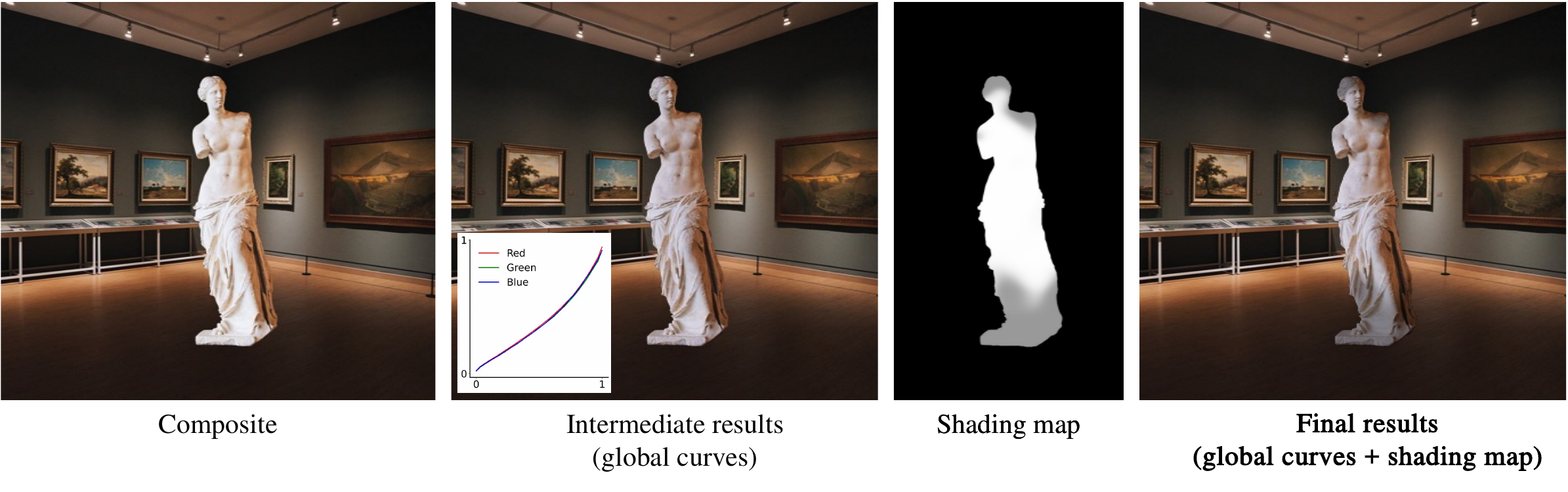}
  \vspace{-.15in}
   \caption{\label{fig:intermediate}\textbf{Intermediate results and parametric outputs.} RGB curves harmonize the global color/tone (center), while our shading map corrects the local shading in the harmonization output (right).}
\end{figure}

To better visualize the roles of our two-stage parametric model, Figure \ref{fig:intermediate} shows the intermediate results as well as the parametric outputs (global curves and shading map) of a representative example. The global curves module harmonizes the global tone of the foreground sculpture and matches it with the background scene. The shading map module further performs local editing to adjust the shading of the sculpture and fit it with the lighting environment better.

%% file: TexFolder/table_iharmony.tex
\begin{table}[ht]
\centering
\setlength\tabcolsep{1.5pt} 
\begin{tabular}{@{}llllll@{}} 
\toprule
\begin{tabular}[c]{@{}l@{}}Size\end{tabular}                  & Method     & MSE $\downarrow$ & PSNR $\uparrow$ & \begin{tabular}[c]{@{}l@{}}SSIM $\uparrow$\\$\times 10^{-2}$\end{tabular}& \begin{tabular}[c]{@{}l@{}}LPIPS $\downarrow$\\$\times 10^{-3}$\end{tabular}                \\ 
\midrule
\multirow{6}{*}{\begin{tabular}[c]{@{}l@{}}$256$\end{tabular}} & Composite  & 172.3  & 31.74     & 97.48     & 16.46                       \\ 
                                                                     &  DovNet \cite{cong2020dovenet}          &    51.33 & 34.97      &  98.12    &            9.734            \\
                                                                     &  IHT \cite{guo2021image}          &  30.46   &  37.33    &   98.77   &  \blue{\textbf{7.347}}
                                                                        \\
                                                                     &  Harmonizer \cite{ke2022harmonizer}       & \blue{\textbf{24.24}}&  \blue{\textbf{38.25}}    & \red{\textbf{99.09}}     &    7.349                    \\ 
                                                                     & \textbf{Ours} &  \red{\textbf{20.57}}   & \red{\textbf{38.30}}     & \blue{\textbf{98.91}}      &    \red{\textbf{7.270}}                    \\ 
\midrule
\multirow{6}{*}{\begin{tabular}[c]{@{}l@{}}$2048^*$ \\\end{tabular}}   & Composite  & 352.9    &  28.39    &  96.36    &         14.52               \\ 
                                                                     &    DovNet \cite{cong2020dovenet}         &   66.37  & 34.01      &  96.35    &   21.45             \\
                                                                     &   IHT \cite{guo2021image}          &  47.34 &  35.12    &   96.53   &      20.65                  \\
                                                                     & Harmonizer \cite{ke2022harmonizer}           &  \blue{\textbf{23.30}}   &   \red{\textbf{38.33}}   &   \blue{\textbf{98.77}}   &   \blue{\textbf{7.148}} \\ 
                                                                     & \textbf{Ours} & \red{\textbf{20.31}}    &  \blue{\textbf{38.29}}  &  \red{\textbf{98.82}}    &  \red{\textbf{7.123}}                      \\
\bottomrule
\end{tabular}
\vspace{-2mm}
\caption{\label{table:iharmony}\textbf{Quantitative comparison on iHarmony benchmark\cite{cong2020dovenet}} at both $256\times256$ and $2048\times2048$. ($*$) We only calculate the metrics on the Adobe5k dataset (a subset of iHarmony4) for high-resolution images. \red{\textbf{Red}}, and \blue{\textbf{Blue}} correspond to the first and second best results. $\uparrow$ means higher the better, and $\downarrow$ means lower the better.}
\label{table:iharm}
\vspace{-8mm}
\end{table}

%% file: TexFolder/table_cooper_rhm.tex
\vspace{-3mm}

\begin{table}[ht]
\centering
\setlength\tabcolsep{1.5pt} 
\begin{tabular}{@{}llllll@{}} 
\toprule
\begin{tabular}[c]{@{}l@{}}Dataset\end{tabular}                  & Method     & MSE $\downarrow$ & PSNR $\uparrow$ &\begin{tabular}[c]{@{}l@{}}SSIM $\uparrow$\\$\times 10^{-2}$\end{tabular}& \begin{tabular}[c]{@{}l@{}}LPIPS $\downarrow$\\$\times 10^{-3}$\end{tabular}                \\ 
\toprule

\multirow{6}{*}{\begin{tabular}[c]{@{}l@{}}\textit{Artist-}\\\textit{Retouched} \end{tabular}} & Composite  & 603.20  & 23.41     & 91.19     & 40.18                       \\ 
                                                                     &  DovNet \cite{cong2020dovenet}          &    352.4 & 26.42      &  90.83   &            56.47            \\
                                                                     &  IHT \cite{guo2021image}          &  369.3   &  26.36    &   90.87   & 55.80
                                                                        \\
                                                                     &  Harmonizer \cite{ke2022harmonizer}       & \blue{\textbf{239.1}}&  \blue{\textbf{29.42}}    & \blue{\textbf{93.84}}     &    \blue{\textbf{33.75}}                    \\ 
                                                                     & \textbf{Ours} &  \red{\textbf{170.1}}   & \red{\textbf{29.79}}     & \red{\textbf{94.56}}      &    \red{\textbf{29.18}}                    \\ 

\midrule

\multirow{6}{*}{\begin{tabular}[c]{@{}l@{}}RealHM \end{tabular}}  & Composite  & 404.4    &  25.88    &  94.70    &         29.32               \\ 
                                                                     &    DovNet \cite{cong2020dovenet}         &   225.1  & 26.72     &  92.00    &   47.50             \\
                                                                     &   IHT \cite{guo2021image}          &  264.0 &  26.48    &   92.46   &      48.48                  \\
                                                                     & Harmonizer \cite{ke2022harmonizer}           &  \blue{\textbf{231.4}}   &   \blue{\textbf{27.40}}   &   \blue{\textbf{94.86}}   &   \blue{\textbf{27.62}} \\ 
                                                                     & \textbf{Ours} & \red{\textbf{153.3}}    &  \red{\textbf{28.34}}  &  \red{\textbf{95.51}}    &  \red{\textbf{23.09}}                      \\

\bottomrule
\end{tabular}
\label{table:iharmony}
\vspace{-2mm}
\caption{\textbf{Quantitative Comparison on RealHM benchmark and \textit{Artist-Retouched} dataset.} Our approach outperforms other methods in all four metrics.}
\label{table:realhm}
\vspace{-6mm}
\end{table}

%% file: TexFolder/table_user_study.tex
\begin{table}[!ht]
\centering
\begin{tabular}{@{}lr@{}} 
\toprule
Methods    & B-T Score $\uparrow$  \\ 
\midrule
Composite  &   0.1025         \\
\addlinespace
DovNet\cite{cong2020dovenet}    &     0.1342       \\
IHT \cite{guo2021image}       &       0.2350     \\
Harmonizer\cite{ke2022harmonizer} &    0.2257        \\ 
\addlinespace
\textbf{Ours} &           \red{\textbf{0.3025}} \\
\bottomrule
\end{tabular}
\vspace{-2mm}
\caption{\textbf{User Study Results.} B-T scores of composite image, DovNet\cite{cong2020dovenet}, IHT\cite{guo2021image}, Harmonizer\cite{ke2022harmonizer} are calculated on 100 real composite images. Our approach ranks first, suggesting superior real-world performance.}
\label{table:btscore}
\vspace{-2mm}
\end{table}

%% file: TexFolder/Conclusion.tex
\section{Conclusion}
In this work, we propose a novel semi-supervised dual-stream training strategy to bridge the training-testing domain gap and mitigate the generalization issues that limit previous works for real-world image harmonization. Our method leverages high-quality artist-created image pairs and unpaired realistic composites to enable richer image edits for real-world applications. Besides, we introduce a new two-stage parametric model (\textit{Global RGB Curves} and \textit{shading map}) to reap the most benefits from our training strategy and, for the first time, enable local editing effects with learned shading map. Our method outperforms other state-of-the-art methods on established benchmarks and real composites. Furthermore, our training strategy has the potential to generalize to a wider range of image harmonization operations (e.g., matching the noise, harmonizing the boundaries, adding cast shadows). As a future work, we would like to include more attributes in our models and further improve the performance of real-world image harmonization.

%% file: TexFolder/Supplementary_og.tex
\section{Supplementary material}

In this supplementary, we first describe how our \textit{Artist-Retouched} dataset was constructed, and show representative visual examples from the dataset (introduced in \S~\ref{section:dual}).
Then, we show the visual comparisons on iHarmony dataset~\cite{cong2020dovenet} and report the detailed quantitative comparison results on four subsets (HCOCO, HAdobe5k, HFlickr, Hday2night).
Besides, we provide more high-resolution visual results of different methods on the \textit{Artist-Retouched} dataset and RealHM benchmark (supplementary to \S~\ref{sec:quant}).
We then go in-depth into our real composite dataset with captured references and present more qualitative visual comparisons (supplementary to \S~\ref{section:user_study}). Furthermore, we show more visual comparisons of real composite images we used as part of our user studies (\S~\ref{section:user_study}).
Finally, we show more intermediate results and parametric outputs of our method for real-composite image harmonization (supplementary to \S~\ref{sec:abl}).
%

\subsection{\textbf{\textit{Artist-Retouched}} dataset}

In this work, we propose to use a new \textit{Artist-Retouched} dataset for our dual-stream training experiments.
Unlike previous work, \textit{Artist-Retouched} contains image pairs retouched by artists rather than mostly relying on random color augmentations.
Artists were allowed to use global luminosity or color adjustments operations, but also local editing tools like brushes, e.g., to alter the shading.
All the image editing was done using Adobe Lightroom, a software dedicated to photo adjustment.
Figure \ref{fig:s1} shows representative before-after image pairs in the \textit{Artist-Retouched} dataset.
\textit{Artist-Retouched} consists of $n=46173$ before/after retouching image pairs $\{\mathbf{I}_i, \mathbf{O}_i\}_{i=1,\ldots,n}$, with the foreground mask $\mathbf{M}_i$ for each pair.
As visualized in the figure, the retouching procedure consists of global luminosity/color adjustments (e.g., exposure, contrast, Highlights, Temp, Tint, Hue) and local editing tools(e.g., adding shading, creating soft transitions by gradient mask).
From each triplet $\{\mathbf{I}_i, \mathbf{O}_i, \mathbf{M}_i\}$, we can generate two synthetic composites inputs for training: one with only the foreground retouched $\mathbf{M}_i\cdot\mathbf{O}_i + (1-\mathbf{M}_i)\cdot\mathbf{I}_i$, and the other one with only the background being retouched $\mathbf{M}_i\cdot\mathbf{I}_i + (1-\mathbf{M}_i)\cdot\mathbf{O}_i$.
We use the unedited image $\mathbf{I}_i$ and the retouched image $\mathbf{O}_i$ as ground truth targets of these composite inputs, respectively.

\subsection{More results on iHarmony benchmark}

As discussed in \S\ref{sec:quant}, we evaluate our method on the iHarmony benchmark~\cite{cong2020dovenet} and present the quantitative results on the entire dataset.
In this section, we report the quantitative results on four subsets of iHarmony --- HCOCO, HAdobe5k, HFlickr, Hday2night.
We compare our method with DovNet\cite{cong2020dovenet}, IHT\cite{guo2021image}, Harmonizer\cite{ke2022harmonizer}. Our method outperforms or matches state-of-the-art approaches in all four subsets of iHarmony benchmark.
Table \ref{sub:table1} summarizes the quantitative results.
Besides, Figure \ref{fig:s2} shows a gallery of selective visual comparisons between different approaches at $512\times 512$ resolution. 
For better visualization, we resize the images to their original aspect ratios.

\subsection{More visual results on \textbf{\textit{Artist-retouched}} dataset}

As introduced in \S~\ref{section:dual}, we evaluate different methods on a testing split of our \textit{Artist-Retouched} dataset with realistic retouches from human experts. In addition to the results shown in Figure~\ref{fig:visual_comparison}, Figure~\ref{fig:s3} presents more visual comparisons on  \textit{Artist-Retouched} testing dataset at $1024$ resolution.
We observe that our results agree better with the ground truth images in terms of the visual quality compared to other methods (DovNet~\cite{cong2020dovenet},IHT~\cite{guo2021image},Harmonizer~\cite{ke2022harmonizer}).
Besides, we also show one failure example (boat), where all methods (including ours) fail to retrieve the correct color of the ground truth image, though some of them look harmonious by themselves without seeing the reference.
We hypothesize that, in this case, the skylight illumination in the ground-truth image is difficult to infer from the background.

\subsection{More visual results on \textbf{\textit{RealHM}} benchmark}

Different from synthetic dataset~\cite{cong2020dovenet}, RealHM~\cite{jiang2021ssh} benchmark contains 216 real-world high-resolution composites with expert annotated harmonization results as ground truth.
In this section, we present more visual harmonization comparisons in Figure~\ref{fig:s4} at $1024$ resolution.
From the results, we observe that even though there exist strong foreground/background color mismatches in the real composite images, our method produces more harmonious results compared to other approaches.

\subsection{Real composites with captured reference}

As briefly introduced in \S~\ref{section:user_study}, for qualitative evaluation, we created a dataset of 40 high-resolution real-composite images with captured references.
As illustrated in Figure \ref{fig:s5}, we first capture a fixed set of foreground objects against multiple backgrounds (scenes), as well as the corresponding "background-only" images.
We then segment the foreground object of one photo and paste it onto the "background-only" photo of another with roughly the same location.
The captured photo of the same object in the same background scene serves as a reference for qualitative evaluation. Figure \ref{fig:s6} visualize selective examples of the harmonization results.
We compare our method with state-of-the-art approaches. As shown in the figure, for the first example, our result shows better visual agreements with the captured reference.
For the second and third examples, though our results don't exactly match the reference (none of the other methods does), our method still produces images with harmonious appearances.
We will release this dataset upon publication.

\subsection{More results on real composite images}

Figure~\ref{fig:s7} shows more visual comparisons on real composite images where we don't have the ground truth or captured reference. We use these images as part of our user studies (\S\ref{section:user_study}). We compare our method with DovNet~\cite{cong2020dovenet}, IHT~\cite{guo2021image}, Harmonizer~\cite{ke2022harmonizer}. We will release these testing real composite images upon the publication of this work.

\subsection{More intermediate results}
Figure~\ref{fig:s8} presents more intermediate results and parametric outputs on RealHM\cite{jiang2021ssh} real-composite benchmark.
As shown in the results, our predicted RGB curves harmonize the global color/tone, while our learned shading map incorporates local shading to the final outputs.
By comparing with the human-annotated harmonization results (right), we observe that our local shading maps align well with the local operation done by the human experts.
For instance, for the top three rows, both our results and the human-annotated ground truth selectively darkened the bottom part of the foreground objects.
For the fourth row example, our result highlights the region with incoming light while darkening other foreground parts, which agrees with the operations done by human experts.

\subsection{Demo video}

To further better demonstrate the effectiveness of our method in real-world applications, we prepared and recorded a demo video (see attachments of the supplementary material). We can interactively run our demo on a single CPU without access to extensive computing resources.

\renewcommand{\thefigure}{S1}
\begin{figure*}[ht]
  \centering
   \includegraphics[width=1\linewidth]{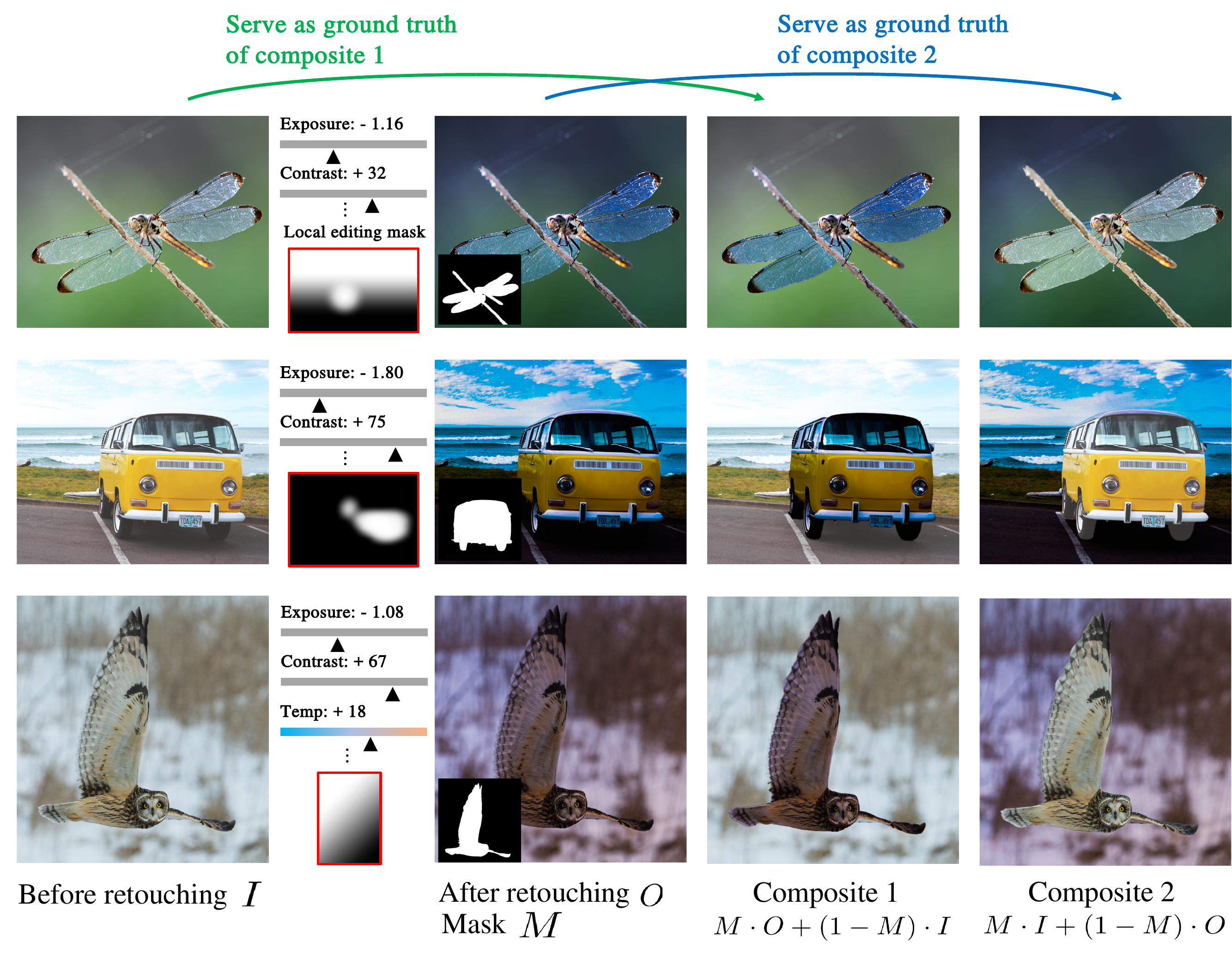}
   \caption{\label{fig:s1}\textbf{Construction of \textit{Artist-Retouched} dataset.} \textit{Artist-Retouched} dataset contains before/after artist-retouching image pairs  $\{\mathbf{I}_i, \mathbf{O}_i\}_{i=1,\ldots,n}$ with the foreground mask $\mathbf{M}_i$ for each pair. Artist retouching procedures include both global luminosity/color adjustments as well as local editing. Local editing masks (images with red borders) in the figure indicate the selective regions where artists perform local editing (e.g., shading). Two composite images (Composite 1 and 2) are created and used for training from each pair of images. Unedited image $I$ and retouched image $O$ serve as the ground truth for composite 1 and 2, respectively.
    }
\end{figure*}

\renewcommand{\thetable}{S1}
\begin{table*}[ht]
\centering
\setlength\tabcolsep{4pt} 
\begin{tabular}{ccccccccccc} 
\toprule

\multirow{2}{*}{Method} & \multicolumn{2}{c}{HCOCO} & \multicolumn{2}{c}{Adobe5k} & \multicolumn{2}{c}{HFlickr} & \multicolumn{2}{c}{Hday2night} & \multicolumn{2}{c}{Entire dataset}  \\
                        & PSNR$\uparrow$ & SSIM $\uparrow$              & PSNR$\uparrow$ & SSIM $\uparrow$                & PSNR$\uparrow$ & SSIM$\uparrow$                 & PSNR$\uparrow$ & SSIM $\uparrow$                   & PSNR$\uparrow$ & SSIM $\uparrow$                        \\ 
\midrule
Composite               &   33.92   &     0.9862               &  28.51    &     0.9563                 &   28.44   &     0.9638                &  34.32    &      0.9741                   &   31.74   &     0.9748                         \\
DovNet\cite{cong2020dovenet}                  &    35.76  &   0.9875                 &   35.05   &    0.9733                  &    30.68  &          0.9711             &  34.83    &          0.9707               &  34.97    &      0.9812                       \\
IHT\cite{guo2021image}                     &  38.38    &     0.9924               &   37.02   &      0.9819                &  32.84    &        \textbf{\blue{0.9810}}              &   36.79   &      0.9763                   &  37.33     &     0.9877                         \\
Harmonizer\cite{ke2022harmonizer}              &   \textbf{\blue{38.77}}   &      \textbf{\blue{0.9936}}              &   \textbf{\red{38.97}}   &    \textbf{\red{0.9888}}                  &     \textbf{\red{33.71}} &     \textbf{\red{ 0.9833}}                &    \textbf{\blue{37.96}}  &           \textbf{\blue{0.9813}}              & \textbf{\blue{38.25}}     &           \textbf{\red{0.9909}}                   \\
Ours                 &   \textbf{\red{39.07}}    &     \textbf{\red{0.9940}}               & \textbf{\blue{38.53}}     &    \textbf{\blue{0.9835}}                  & \textbf{\blue{33.60}}     &  0.9793                    &  \textbf{\red{38.15}}    &   \textbf{\red{0.9817}}    &    \textbf{\red{38.30}}  &     \textbf{\blue{0.9891}}                         \\
\bottomrule
\end{tabular}
\caption{\label{sub:table1}\textbf{Quantitative comparisons on subsets of iHarmony benchmark\cite{cong2020dovenet}} at $256\times256$ resolution. We compare our method with DovNet\cite{cong2020dovenet}, IHT\cite{guo2021image}, Harmonizer\cite{ke2022harmonizer}. PSNR and SSIM are used as metrics. \red{\textbf{Red}}, and \blue{\textbf{Blue}} correspond to the first and second best results. $\uparrow$ means higher the better, and $\downarrow$ means lower the better.}
\end{table*}

\renewcommand{\thefigure}{S2}
\begin{figure*}[ht]
  \centering
   \includegraphics[width=1\linewidth]{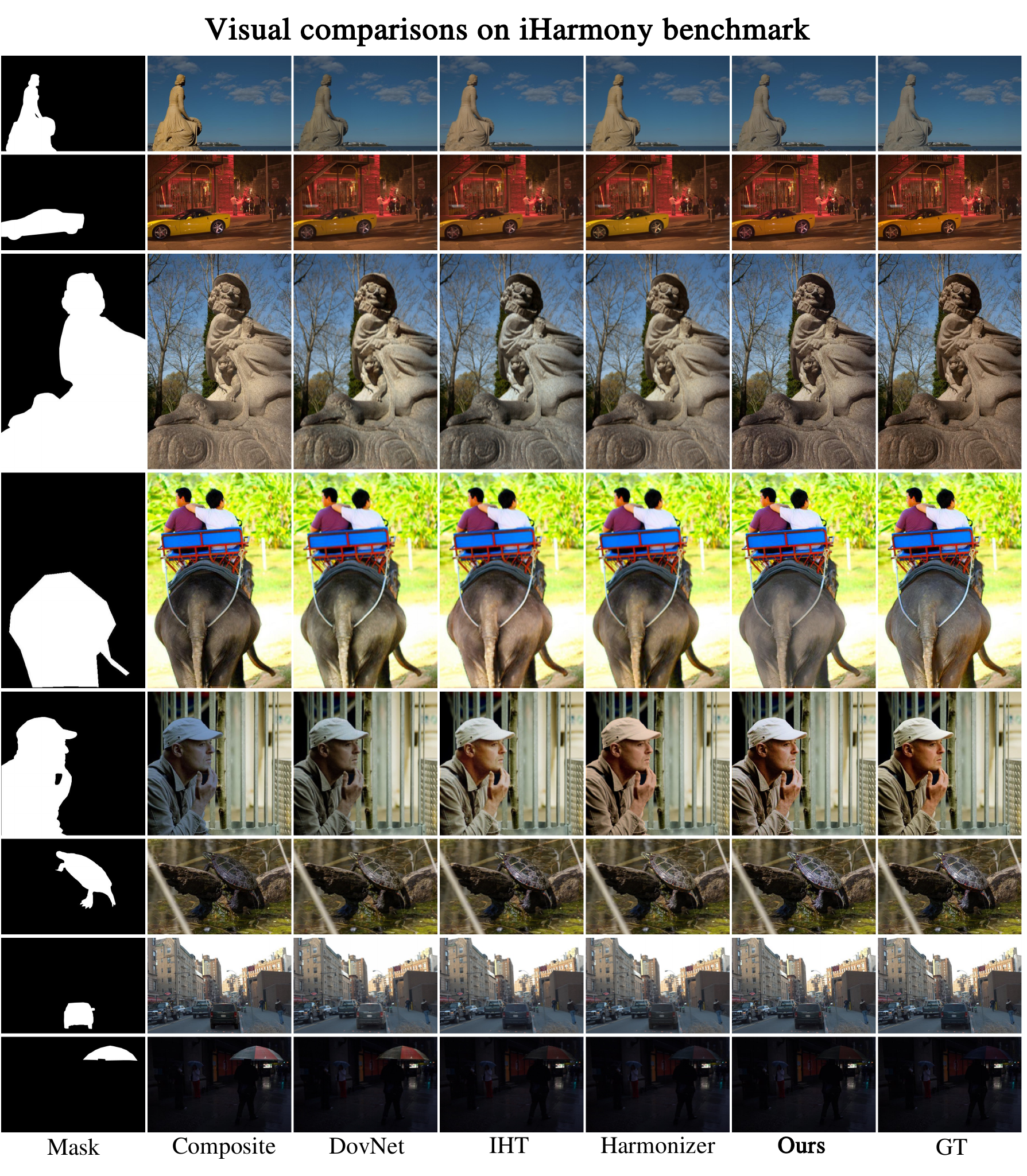}
   \caption{\label{fig:s2}\textbf{Representative visual comparisons between state-of-the-art harmonization methods on iHarmony benchmark.} We compare our method with composite image, DovNet~\cite{cong2020dovenet}, IHT~\cite{guo2021image}, Harmonizer~\cite{ke2022harmonizer}, and ground truth. Foreground masks are displayed in the first column. For better visualization, we resize the images to the original aspect ratio. Our method shows better visual alignment with the ground truth images than other state-of-the-art methods.
    }
\end{figure*}

\renewcommand{\thefigure}{S3}
\begin{figure*}[ht]
  \centering
   \includegraphics[width=1\linewidth]{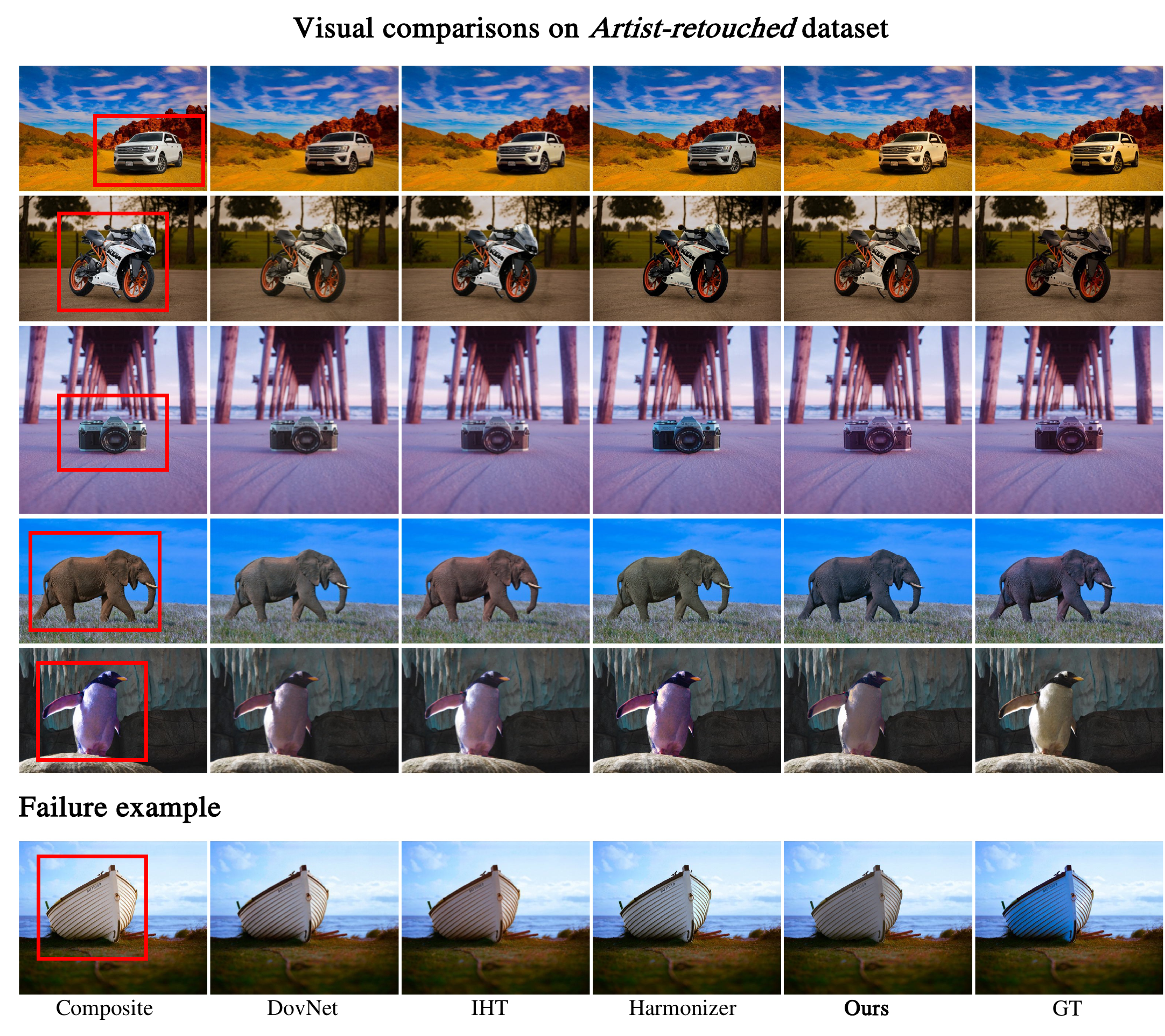}
   \caption{\label{fig:s3}\textbf{More visual comparisons between state-of-the-art harmonization methods on \textbf{\textit{Artist-Retouched}} dataset.} We compare our method with composite image, DovNet~\cite{cong2020dovenet}, IHT~\cite{guo2021image}, Harmonizer~\cite{ke2022harmonizer}, and ground truth. Red boxes indicate the foreground mask of the composite images. Our method shows better visual alignment with the ground truth images compared to other state-of-the-art methods. We also present one failure example, where all methods fail to recover the ground truth appearance, though some of them look harmonious without referring to the ground truth.
    }
\end{figure*}

\renewcommand{\thefigure}{S4}
\begin{figure*}[ht]
  \centering
   \includegraphics[width=1\linewidth]{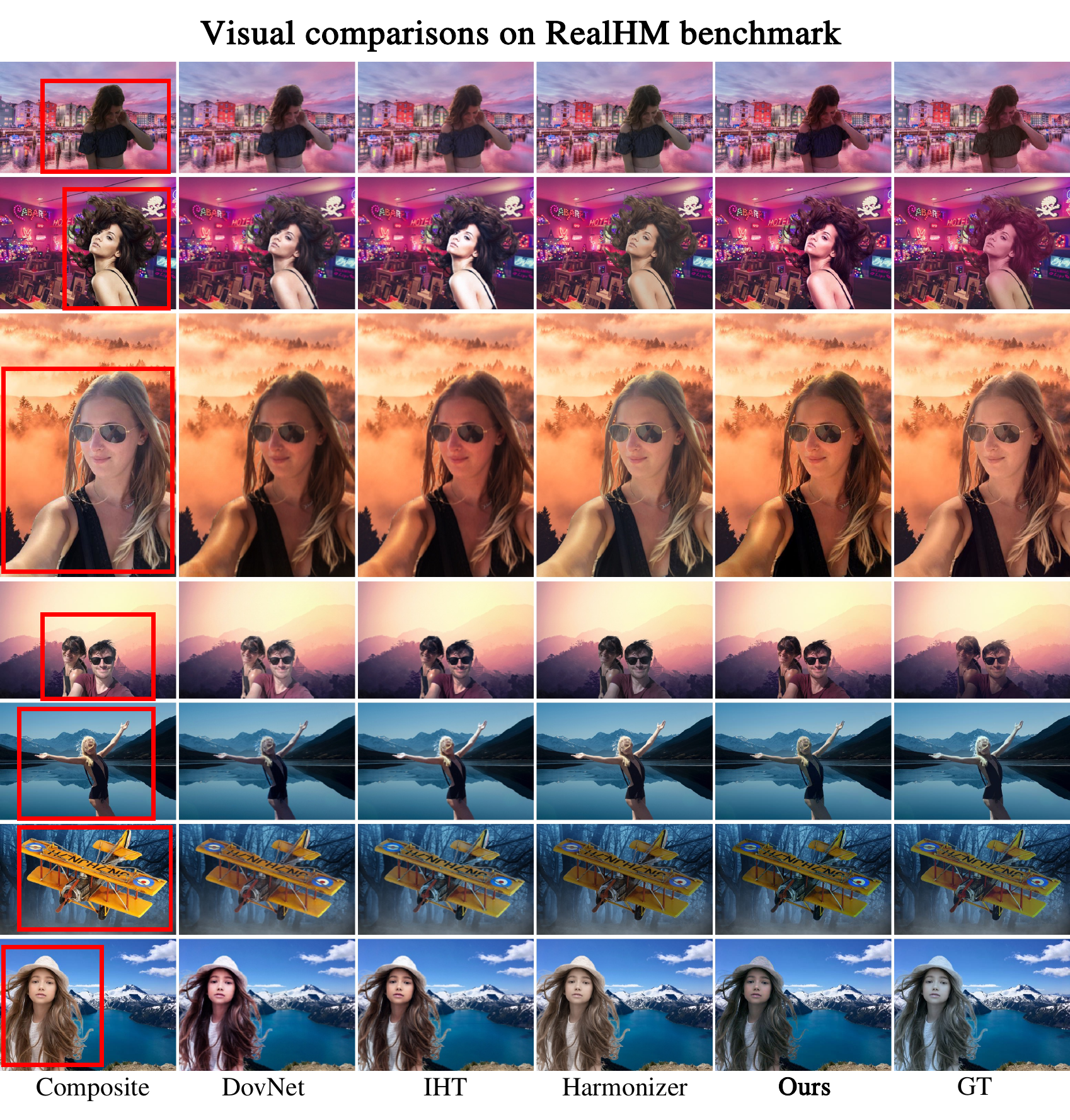}
   \caption{\label{fig:s4}\textbf{More visual comparisons between state-of-the-art harmonization methods on RealHM benchmark.} We compare our method with composite image, DovNet~\cite{cong2020dovenet}, IHT~\cite{guo2021image}, Harmonizer~\cite{ke2022harmonizer}, and ground truth. Our method shows better color consistency with the ground truth images (row 1, 2, 4, 6, and 7) and deliver more harmonious results.
    }
\end{figure*}

\renewcommand{\thefigure}{S5}
\begin{figure*}[ht]
  \centering
   \includegraphics[width=1\linewidth]{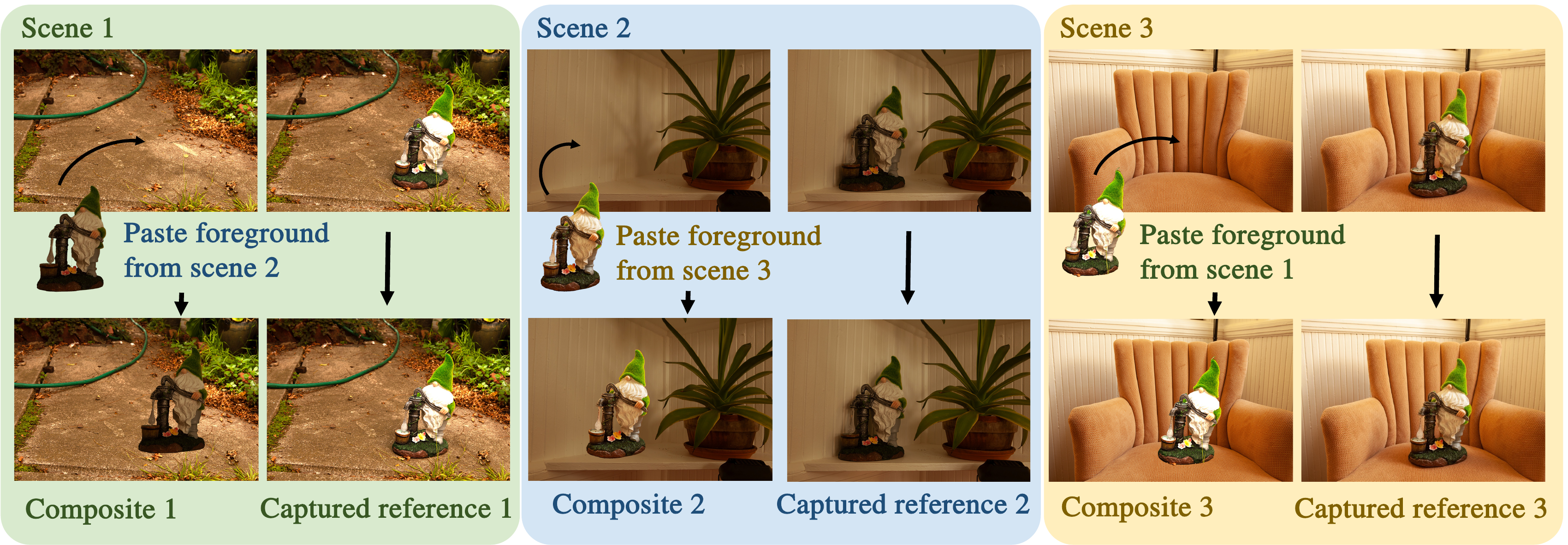}
   \caption{\label{fig:s5}\textbf{Construction of composite images with captured reference.} First, we capture the same foreground object against multiple backgrounds (3 backgrounds in the figure), as well as the corresponding "background-only" photos. We then segment the foreground object from one photo and paste it onto the "background-only" image of another to generate the composite images. The captured photo of the same object in the same background scene serves as qualitative references (Here, captured references 1, 2, and 3).
    }
\end{figure*}

\renewcommand{\thefigure}{S6}
\begin{figure*}[ht]
  \centering
   \includegraphics[width=1\linewidth]{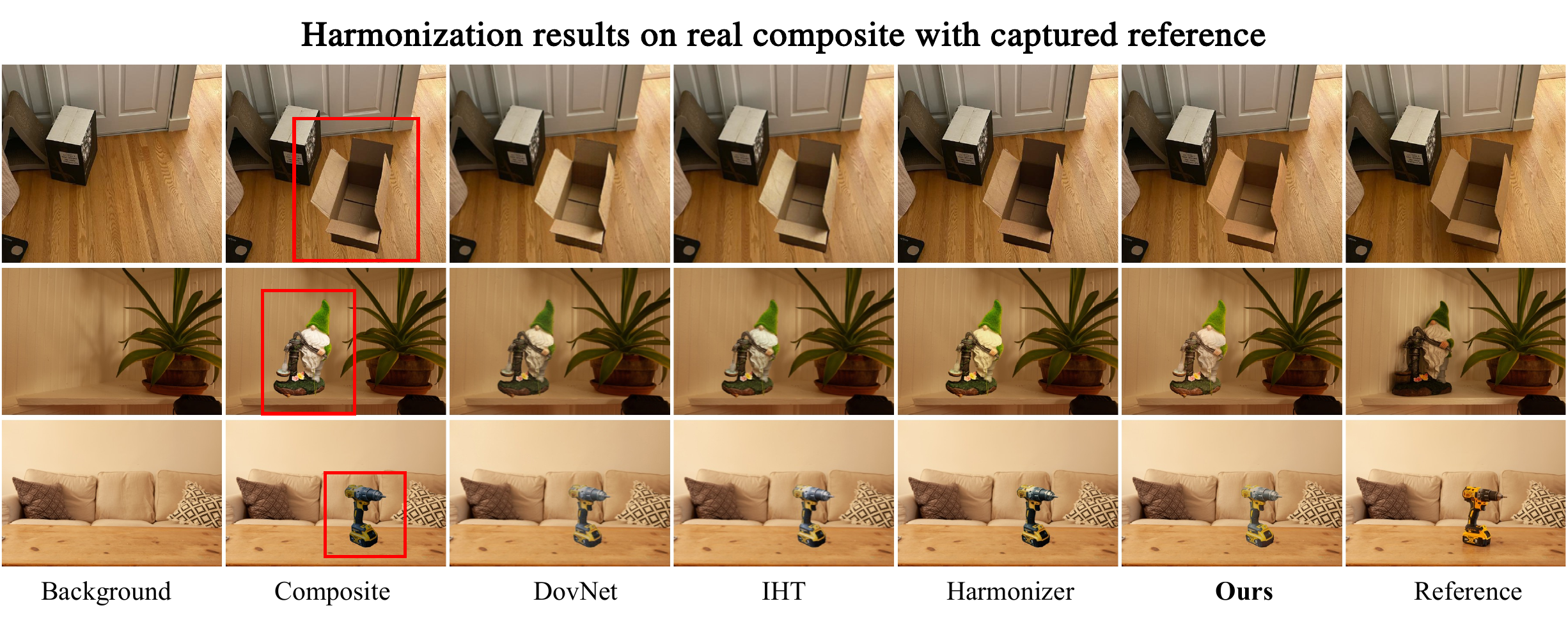}
   \caption{\label{fig:s6}\textbf{Real composite harmonization results with captured reference.} The composite is obtained by pasting the foreground object from a different photo (not shown) onto the background (left). The reference (right) is obtained by physically placing the foreground object in the background scene and taking a photo. We compare our method with composite image, DovNet~\cite{cong2020dovenet}, IHT~\cite{guo2021image}, Harmonizer~\cite{ke2022harmonizer}, and the captured reference.
    }
\end{figure*}

\renewcommand{\thefigure}{S7}
\begin{figure*}[ht]
  \centering
   \includegraphics[width=0.9\linewidth]{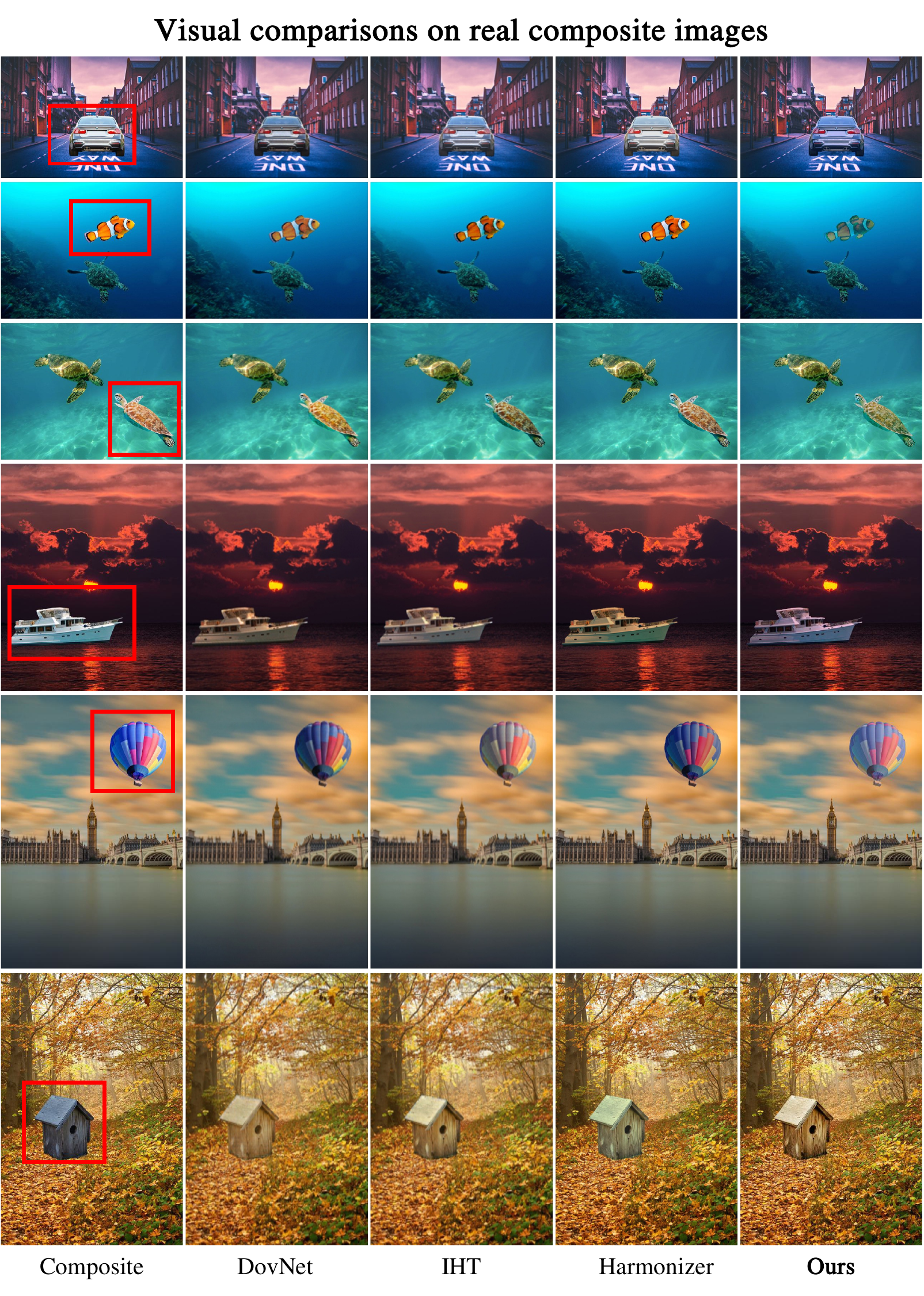}
   \caption{\label{fig:s7}\textbf{More visual comparisons on real composite images.} We compare our method with composite image, DovNet~\cite{cong2020dovenet}, IHT~\cite{guo2021image}, and Harmonizer~\cite{ke2022harmonizer}.
    }
\end{figure*}

\renewcommand{\thefigure}{S8}
\begin{figure*}[ht]
  \centering
   \includegraphics[width=1\linewidth]{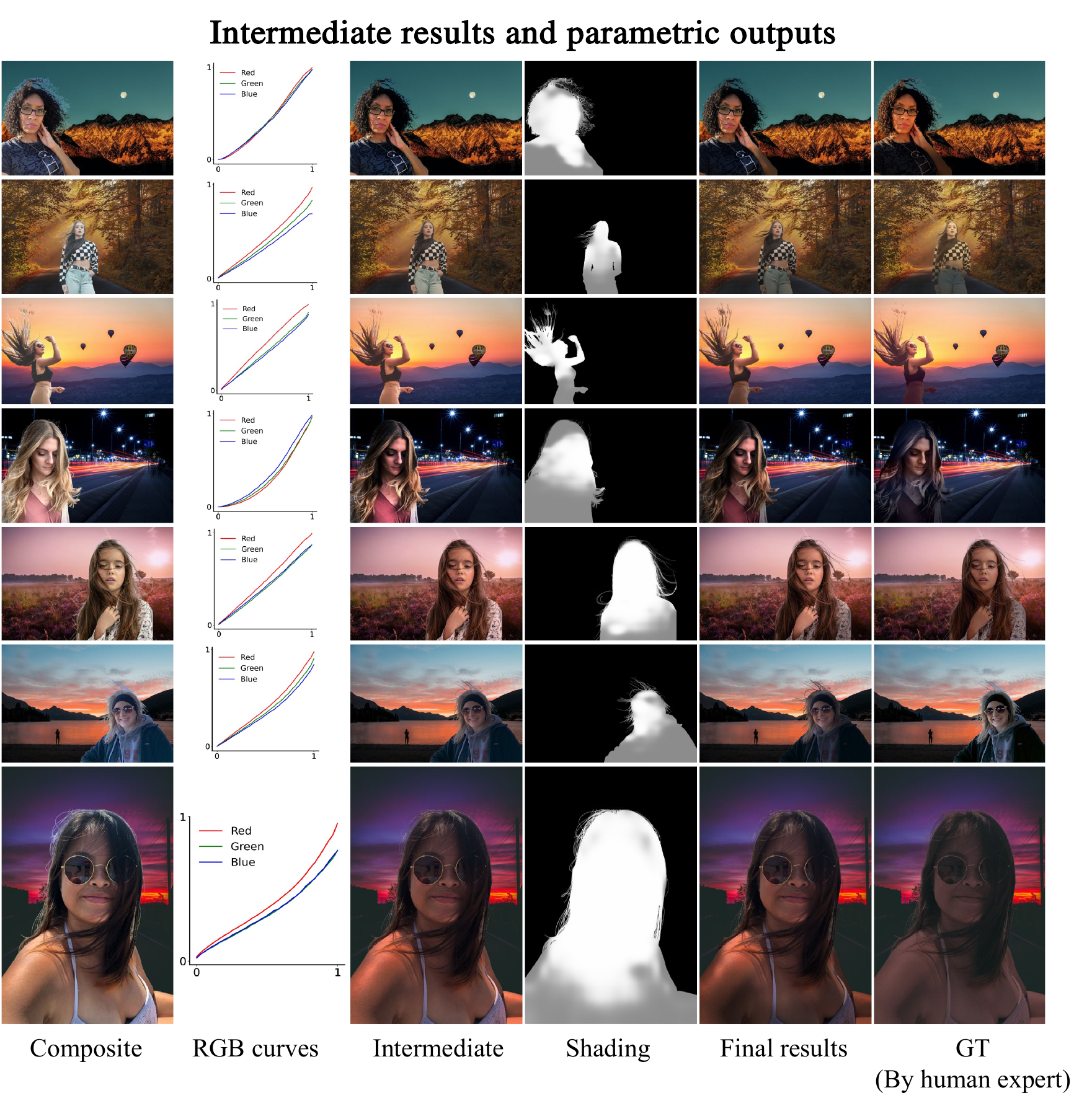}
   \caption{\label{fig:s8}\textbf{Intermediate results and parametric outputs on RealHM benchmark.} RGB curves harmonize the global color/tone (third column), while our shading map corrects the local shading in the final harmonization outputs (fifth column). Our local shading maps agree well with the local shading operations done by human experts/artists (right column).
    }
\end{figure*}